\documentclass[5p]{elsarticle}
\usepackage[titletoc]{appendix}
\usepackage{graphicx}
\usepackage{subfigure}
\usepackage{amsmath,bm}
\usepackage{stfloats}

\journal{Neurocomputing}

\begin{document}
\begin{frontmatter}

\title{See Blue Sky: Deep Image Dehaze Using Paired and Unpaired Training Images}

\author[a]{Xiaoyan Zhang }
\author[a]{Gaoyang Tang }
\author[a]{Yingying Zhu\corref{cor1}}
\cortext[cor1]{Corresponding author}
\ead{zhuyy@szu.edu.cn}
\author[b]{Qi Tian }
\address[a]{Xiaoyan Zhang, Gaoyang Tang and Yingying Zhu are with the College of Computer Science and Software Engineering, Shenzhen University, Shenzhen, Guang Dong 518060, PR China}
\address[b]{Department of Computer Science, The Univerisity of Texas at San Antonio, TX 78249, USA}
\begin{abstract}
The issue of image haze removal has attracted wide attention in recent years. However, most existing haze removal methods cannot restore the scene with clear blue sky, since the color and texture information of the object in the original haze image is insufficient. To remedy this, we propose a cycle generative adversarial network to construct a novel end-to-end image dehaze model. We adopt outdoor image datasets to train our model, which includes a set of real-world unpaired image dataset and a set of paired image dataset to ensure that the generated images are close to the real scene. Based on the cycle structure, our model adds four different kinds of loss function to constrain the effect including adversarial loss, cycle consistency loss, photorealism loss and paired L1 loss. These four constraints can improve the overall quality of such degraded images for better visual appeal and ensure reconstruction of images to keep from distortion. The proposed model could remove the haze of images and also restore the sky of images to be clean and blue (like captured in a sunny weather).
\end{abstract}

\begin{keyword}
Haze removal \sep Sky reconstruction \sep  Generative adversarial network \sep Photorealism loss

\end{keyword}

\end{frontmatter}

single image haze removal problem has been a hot research topic in recent years. Haze removal is of great importance in our daily life and industry photography. Given a single haze image, traditional methods \cite{he2011single,meng2013efficient,Tarel2009Fast} based on atmospheric physical models \cite{narasimhan2002vision} prefer to judge the concentration according to the intensity or texture of haze image pixels. As a result, the haze of the foreground can be almost removed, but a clear sky in the background cannot be obtained. The reason is that the information of texture and color is insufficient especially in the sky region of the haze image. Moreover, the atmospheric scattering is too complicated to be computed using the atmospheric physical model. The airlight is different for different pixels, thus cannot be measured by a global constant. The transmission value that generally describes the scene radiance attenuation is simplified. Thus the current haze removal methods based on atmospheric physical models cannot remove the heavy haze ideally and the dehaze image looks dim (see the dehaze results in Fig. 1(b)). Meanwhile, the color and texture of the sky are not reconstructed after dehazing while easily creating noises. Images with clean and blue sky (in sunny weather) are more preferred to be captured and shared in social media \cite{laffont2014transient}. Therefore, this paper proposes to restore the scene to have a clean and blue sky after dehazing, like captured in a sunny weather. Therefore, our objective is in two folds: remove the haze of the image, and also reconstruct the image sky to be clean and blue based on the available information of original haze sky (as the results in the bottom row shown in Fig. 1).

\begin{figure}[htbp]
\label{fig:1}
\centering
   \includegraphics[width=0.20\linewidth]{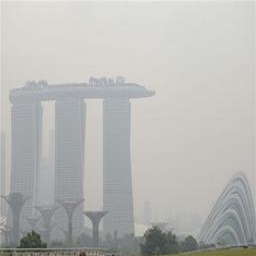}
   \includegraphics[width=0.20\linewidth]{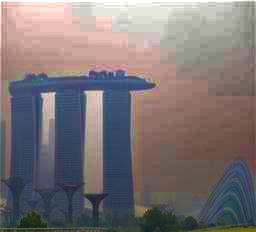}
   \includegraphics[width=0.20\linewidth]{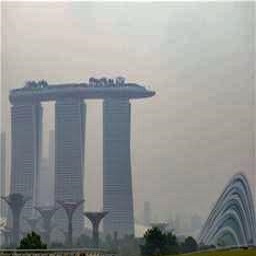}
   \includegraphics[width=0.20\linewidth]{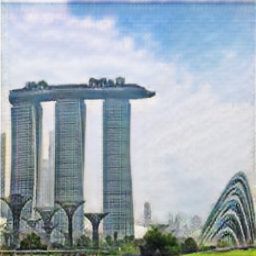}
   \includegraphics[width=0.20\linewidth]{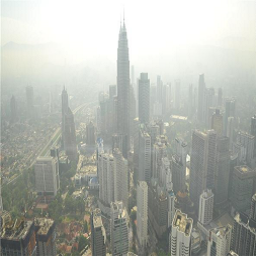}
   \includegraphics[width=0.20\linewidth]{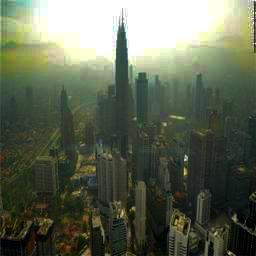}
   \includegraphics[width=0.20\linewidth]{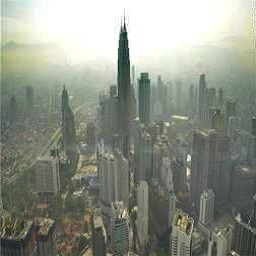}
   \includegraphics[width=0.20\linewidth]{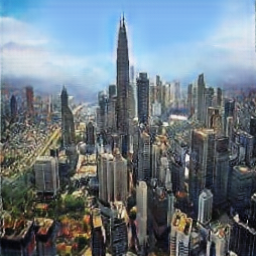}
   \subfigure[]{\includegraphics[width=0.20\linewidth]{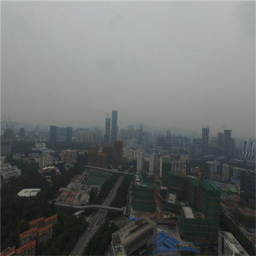}}
   \subfigure[]{\includegraphics[width=0.20\linewidth]{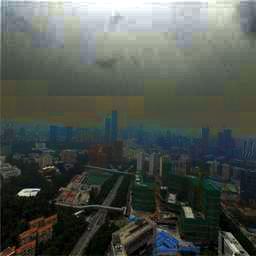}}
   \subfigure[]{\includegraphics[width=0.20\linewidth]{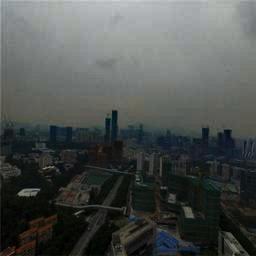}}
   \subfigure[]{\includegraphics[width=0.20\linewidth]{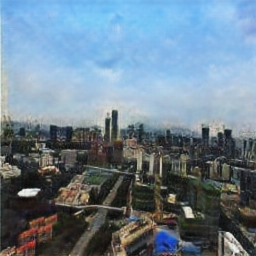}}
   \caption{Some dehaze results. (a)Original images. (b-d)Dehaze results of traditional method of He {\em et al.} \cite{he2011single}, Ren {\em et al.} \cite{ren2016single} and our proposed SCGAN method.}
\label{fig:onecol}
\end{figure}
With the development of neural networks, feature learning from data is much popular. As a representative of neural network models, Generate Adversarial Networks \cite{goodfellow2014generative} (GAN) can even synthesize images. Therefore, this work aims at leveraging this property to train a network that can learn the features of haze-free images with a clear sky and achieve the correct feature transformation for haze removal and sky restoration. However, lack of paired image datasets (composed by haze images and corresponding haze-free images with clean and blue sky) is a bottleneck for our task. Some methods \cite{tang2014investigating, zhu2015fast,cai2016real,ren2016single} use synthesized haze images by supposing that the depth in a patch remains unchangeable in adding haze on haze-free images. The indoor NYU depth dataset \cite{silberman2012indoor} is also used to synthesize haze images. These learning-based methods trained with synthesized haze images still do not work well on restoring the sky region. The sky of results are not clear (see the dehaze results in Fig. 1(c)). CycleGAN \cite{zhu2017unpaired} model allows using unpaired image data to train the model due to its special cycle structure with two generators and two discriminators. The cycle consistency constraint is a benefit for training using the unpaired dataset. It can greatly reduce the difficulty of real paired images collection, as it only needs to collect a set of haze images and a set of clear images from the web seperately. However, in order to accurately learn the profile of objects, training with the paired dataset is better than using only the unpaired dataset. Because the paired dataset can strongly constraint the network to generate the approximate results. In order to take full advantage of paired and unpaired datasets, we propose to train a network using both paired and unpaired datasets for effective haze removal and sky reconstruction.

A Strong Constraint Generative Adversarial Network (SCGAN) model is proposed to learn the feature from two different datasets: unpaired dataset and paired dataset. The unpaired dataset consists of a large number of outdoor haze images and a large number of outdoor clean weather images collected from the Internet. Make3d depth dataset \cite{saxena20083,saxena2009make3d} contains outdoor images with depth information, so we apply it to synthesize paired haze dataset. Our SCGAN model has a cycle structure including two generators and two discriminators, which can be taken as an integration of two GAN models for dehazing and add-hazing, respectively. The input haze images pass through two generators in turn, and the adversarial loss and the cycle consistency loss are calculated to constrain the model to generate the corresponding dehaze images when training with unpaired images. In order to strengthen the content consistency constraint, we also add a photorealism regularization term in the loss function. This constraint is a benefit for the sharpness of the image and correcting the color of objects by using a Laplacian Matting matrix \cite{Levin2008A}. It can successfully avoid the fault of generating fake image edges deviated from training data distribution. The aligned paired dataset constrains the training of the generators by adversarial loss and paired L1 loss. The combination of the two datasets and different kinds of loss items can ensure that the image is reconstructed with visible content and having a clear sky. Some dehaze results generated by our proposed SCGAN method are shown in the bottom row of Fig. 1.

The novelty of our work can be summarized into three aspects:
\begin{itemize}
  \item A novel end-to-end dehaze model is developed with cycle structure which includes two generators and two discriminators. The specifical structure provides the condition to train using different types of datasets and expediently utilizes the multiform loss constraints to learn the feature for dehaze. We use two different outdoor real-world datasets to train our model: an aligned paired image dataset and an unpaired image dataset. They ensure that the generated images are natural anLd clear.
 \item Based on the cycle structure, our model adds four different kinds of loss functions to constrain the effect including adversarial loss, cycle consistency loss, photorealism loss and paired L1 loss. These four loss constraints can improve the overall quality of such degraded images for better visual appeal and ensure enhanced restrict on reconstruction of images from distorting.
 \end{itemize}

The rest of the paper is organized as follows. Section II surveys the related work. Section III presents the details of the proposed SCGAN method. The implementation details of the SCGAN method including datasets, training details and network architecture are introduced in Section IV. Section V presents experimental results and analysis to illustrate the effectiveness of the proposed method. The limitation of the SCGAN method is discussed in Section VI. Finally, Section VII presents the conclusions.

section{Related Work}
\subsection{Haze Removal based on Atmospheric Physical Model}
Haze removal has been a popular problem in recent years, most of the haze removal methods are based on the atmospheric physical models. Nayer and Narasimhan  \cite{Nayar1999Vision,narasimhan2002vision,Narasimhan2003Contrast} had more detailed representation and derivation of atmospheric physical models. They studied the influence of ambient light on image contrast by studying the scattering of light from the haze, and used to inverse the original image of the scene. Based on their model, four different ways were developed to estimate transmission map, which were by concentration coefficients and reflectivity \cite{Fattal2008Single}, by dark channel prior \cite{he2011single}, by combining contextual regularized weighted L1-norm and the inherent boundary constraint \cite{meng2013efficient}, by linear transformation \cite{wang2017fast} and by non-local haze-line \cite{berman2016non}. Tan {\em et al.} \cite{tan2008visibility} restored haze image by modeling Markov random field and enhanced the contrast of image. Tarel {\em et al.} \cite{Tarel2009Fast} proposed to use medium filter to estimate the transmission coefficient instead of minimum filter in dark channel prior method. All of the above methods have different degrees of dehaze ability. However, all of them had a same problem that they didn't focus on the processing of the sky. Their results normally have noise in the sky after dehazing for images under the heavy haze. Some researchers solved this problem by processing the sky region and non-sky region separately based on segmentation. The segmentation of the sky region was performed based on gradient of intensity \cite{zhu2015fast,yu2016image,wang2013single}, transmission map \cite{zhu2017haze,Shi2014Single}, or context-adaptive labels produced by superpixel \cite{Yoon2015Wavelength}. However, it was hard to segment the sky region accurately under the heavy haze condition. And the above methods can't restore a dim heavy haze sky into a natural clear and blue sky.

\subsection{Haze Removal by Machine Learning}
In recent years, more and more learning based haze removal algorithms were proposed. Tang {\em et al.} \cite{tang2014investigating} proposed to combine four types of haze-relevant features with Random Forest to learn a regression model for estimating
the transmission map for hazy images. Cai {\em et al.} \cite{cai2016dehazenet} proposed a DehazeNet model which was the first End-to-End Convolutional Neural Network (CNN) framework for estimating the transmission map. Ren {\em et al.} \cite{ren2016single} proposed a multi-scale CNN model to estimate the transmission map. A ranking CNN model was proposed in \cite{song2018single} to estimate the transmission map. These methods mainly focused on estimating the transmission map. The atmospheric light was calculated separately, and then they used the physical model to recover clear images. In contrast, Li {\em et al.} \cite{li2017aod} proposed an all-in-one dehazing network, where the transmission map and the atmospheric light were estimated in one unified model. GAN was used in haze removal firstly in \cite{zhang2017image}. It performed a joint learning of transmission map and image dehazing. Similarly, Zhang {\em et al.} \cite{zhang2018densely} proposed a Densely Connected Pyramid Dehazing Network (DCPDN) to jointly learn the transmission map, atmospheric light and dehazing all together. In this framework, a joint discriminator-based GAN was used to exploit the structural relationship between the transmission map and the dehazed image. Instead of estimating the physical parameters, Li {\em et al.} \cite{tang2018GAN} proposed to directly generate the haze-free images using conditional GAN.

Although these methods can obtain encouraging results, they are restricted in practical application because they require a large amount of 'paired' data for supervised training of their models. Synthesized hazy images using the depth meta-data from an indoor dataset \cite{silberman2012indoor} or local patches are used to train their models. However, the colors and texture patterns appeared in indoor images and local patches only take a small portion of the natural visual world, which may be insufficient to learning discriminative features for dehazing of real-world images. To avoid the limitation of paired data, Yang {\em et al.} \cite{yang2018towards} proposed to restore haze-free images using only unpaired supervision. They introduced Disentangled Dehazing Network (DDN) to estimate the scene radiance, transmission map, and global atmosphere light by utilizing three generators jointly in the framework of GAN. This model can be trained using the outdoor unpaired images. However, the DDN model was based on the atmospheric physical model and need to train three generators at the training phase. Differently, our main purpose is building an end-to-end network regardless of atmospheric scattering model for single image dehazing.

\subsection{General Adversarial Networks}
Since Ian Goodfellow proposed GAN \cite{goodfellow2014generative} in 2014, it has been widely concerned. GAN includes a generative model (generator) and a discriminative model (discriminator). According to the study, the samples produced by GAN are sharp and clear. However, GAN is difficult to train as the generator and the discriminator are mutually adversarial and it also could produce artifacts in the synthesized images. DCGANs \cite{radford2015unsupervised} added architectural constraints on GAN, and learned a hierarchy of representations from object parts to scenes in both the generator and the discriminator. SimGAN \cite{shrivastava2016learning} model was trained to simulate the real texture of images by two datasets, an unlabeled real dataset and a synthetic image dataset. It was trained by minimizing the combination of a local adversarial loss and a self-regularization term. Pix2pixGAN was not application-specific \cite{isola2016image}, and it added a L1 conditional loss function based on paired dataset. By contrast, there were many generative application-specific models based on conditional GAN, such as inpainting \cite{pathak2016context,iizuka2017globally,Wang2018Video}, super-resolution \cite{ledig2016photo}, colorization \cite{cao2017unsupervised}, style transfer \cite{chidambaram2017style}, age predicition \cite{antipov2017face}, de-raining \cite{He2017Image}, etc. To make up for the shortage of aligned paired dataset, some researches tried to apply forward-backward consistency in the field of image transfer. In the past year, CycleGAN \cite{zhu2017unpaired}, Duel GAN \cite{yi2017dualgan} and DiscoGAN \cite{kim2017learning} show up. All these models used cycle consistency loss to train the GAN network.

\subsection{Photorealism Regularization}
He {\em et al.} \cite{he2011single} adopted Laplacian Matting of Levin {\em et al.} \cite{Levin2008A} as the soft-matting method to estimate the accurate transmission map. The input coarse transmission map was refined based on boundary information of the original image. Photo style transfer method \cite{luan2017deep} extracted locally affine functions with Laplacian Matting, then fine-tuned the stylized image to fit these affine functions. The results of the stylized images are photorealistic and retain the detailed structures of the content image effectively. Guided filter \cite{Kaiming2013Guided} is an adaptive weighted filter, which can transfer the structures of the guidance image to the filtering output fastly. Deep guided filter was also proposed \cite{Wu2018} to generate full-resolution, edge-preserving outputs with low computational cost. The guided filter uses a window to scan the density characteristics of the guided map, so there will be halo in the boundary area when the difference between the two regions in the window is too large. Our work adopts the Laplacian Matting as a photorealism regularization term to limit our network, so that the optimized function can get realistic effect that is close to the real world image.

\begin{figure*}
\centering
\includegraphics[width=0.9\linewidth]{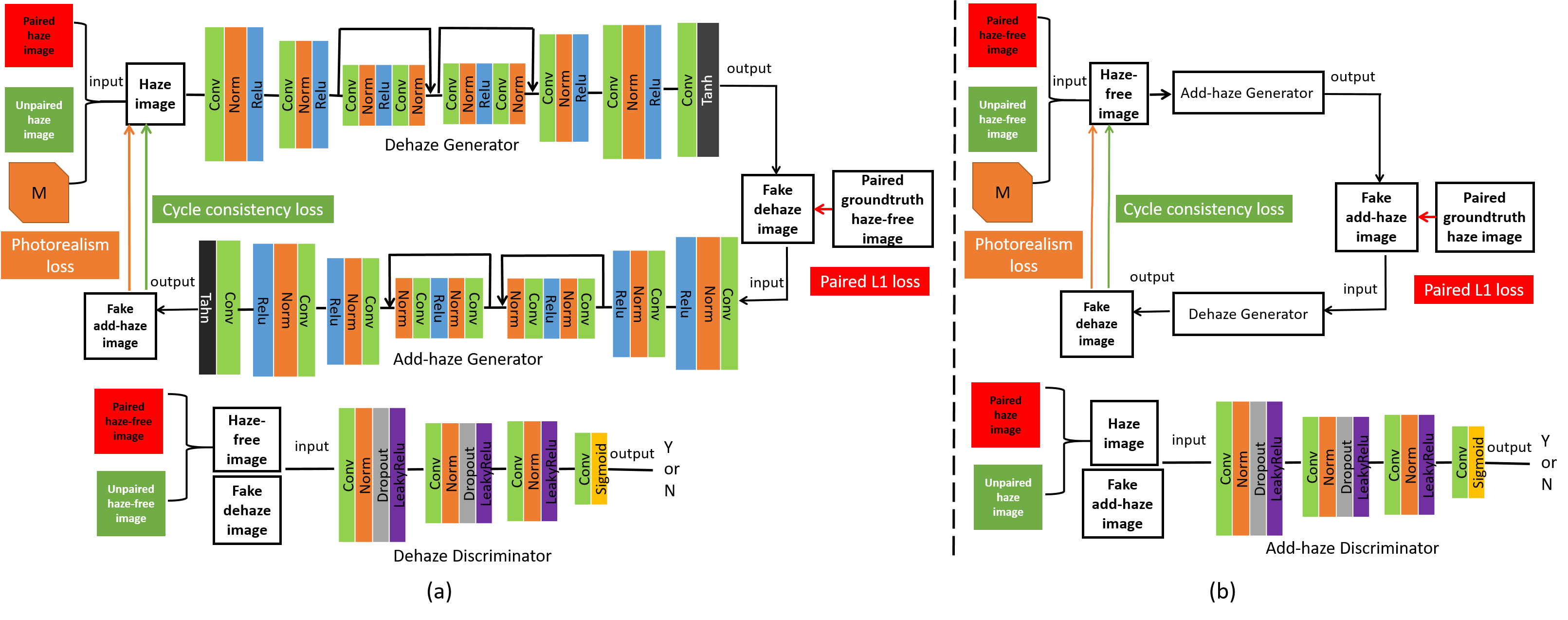}
   \caption{Flowchart of proposed SCGAN method. Our SCGAN model contains two generators and two discriminators. Adversarial Loss is applied to constraint the generators and discriminators using both unpaired dataset and paired dataset. (a) is the forward constraint cycle and (b) is the backward constraint cycle. In the forward constraint cycle, a haze image of the unpaired dataset gets a dehaze image through a dehaze generator, and then it is passed through an add-haze generator to obtain a fake haze image. The fake haze image and original haze image are compared to calculate the cycle consistency loss (in green) and photorealism loss (in orange). The photorealism loss is a regularization term to avoid generating fake image edges by penalizing the fake add-haze image using a locally affine transform carried by \textbf{M} matrix. The \textbf{M} is the image information matrix including the spatial structural information of the original image. Additionally, the dehaze generator is also constrained by the paired images. The paired L1 loss (in red) will be measured between fake dehaze images and ground-truth haze-free images. The dehaze discriminator is trained by using haze-free images from the two datasets and fake dehaze images from the dehaze generator as inputs, and is constrained by adversarial loss. In the backward constraint cycle, for a haze-free image as input, it goes through the add-haze generator and the output image is passed into the dehaze generator to get the fake haze-free image. In a similar manner, the adversarial loss, the cycle consistency loss, the photorealism loss and the paired L1 loss are calculated. The add-haze discriminator has the same structure with the dehaze discriminator and is trained using haze images from the two datasets and fake add-haze images from the add-haze generator as inputs. The detailed network architecture is introduced in Section \ref{sbsec:c}.}
\end{figure*}

\section{Background}
In computer vision and image processing, the widely used image degradation model is as following as \cite{narasimhan2002vision}:
\begin{equation}
\begin{aligned}
I(x) = J(x)t(x) + A(1 - t(x))
\end{aligned}
\end{equation}
where $I$ represents haze image, $J$ is the clear image to be recovered, $A$ is atmospheric light and $t$ is the transmission that is the light which survives the path between the camera and the surface of the objects. It is related to atmospheric degradation factors and the distance of radiation pass through. Therefor, the transmission can be expressed as:
\begin{equation}
\begin{aligned}
t(x) = {e^{ - \beta d(x)}}
\end{aligned}
\end{equation}
where $d$ is the depth of scene, and $\beta$ is the scattering coefficient. Haze-removal from a single image is thus an ill-posed problem, because it requires knowledge of the scene depth d(x), the haze scattering efficient  $\beta$ , and also the atmospheric light. With the development of neural networks, the mapping from the haze image I(x) to the haze-free image J(x) could  be directly learned by a deep learning model, such as
\begin{equation}
\begin{aligned}
{\rm{J(x) =  f(I(x))}}
\end{aligned}
\end{equation}
where $f$ is the mapping function learned by the deep learning model. However, lack of paired image datasets is a bottleneck for designing the deep learning model. Some methods \cite{tang2014investigating},
\cite{zhu2015fast}, \cite{cai2016real}, \cite{ren2016single} use synthesized haze images by supposing that the depth in a patch remains unchangeable in adding haze on haze-free images. The drawbacks of the synthesized data are that the sky of add-haze images is unreal and too monotonous.

These learning-based methods trained with synthesized haze images still do not work well on restoring the sky region. Fortunately, the Internet have an unlimited number of haze-free images with clear sky, which could provide information for attributes of haze removal. Therefore, we proposed to leaning a haze remove model by using paired and unpaired images, which could be modeled as
\begin{equation}
\begin{aligned}
{\rm{J(x) = g(I(x)) =  f(I(x)) + e(I(x))}}
\end{aligned}
\end{equation}
where $f$ is the mapping function learned based on the paired images and $e$ is the learned mapping from unpaired images. They are learned together as a function $g$ by CycleGAN. We refer to as unpaired data.
In this study, the MR-GAN framework is proposed for obtaining accurate synthetic results (estimating MR from CT) simultaneously using the limited paired data and substantial unpaired data. The MR-GAN has two structures—the paired cycle-consistent and unpaired cycle-consistent, to simultaneously train different data.
\section{Proposed SCGAN Method}
Unpaired dataset and paired dataset are used to train our SCGAN model. The flowchart of the proposed method is shown in Fig. 2. Our SCGAN model has a cycle structure including two generators and two discriminators. The two generators are dehaze generator and add-haze generator, and the two discriminators are dehaze discriminator and add-haze discriminator. Adversarial Loss is applied to constraint the generators and discriminators using both unpaired dataset and paired dataset. In the forward constraint cycle based on the unpaired dataset, putting the original haze image into a dehaze generator, and it can produce the dehaze image. Then the dehaze image goes through the add-haze generator to gain the fake add-haze image, which is quite similar to the original input image. The cycle constraint mechanism will calculate the cycle consistency loss between the original haze image and the fake add-haze image (in green color in Fig. 2). Besides, the photorealism loss is calculated as a regularization term (in orange color in Fig. 2) to avoid generating fake image edges by penalizing the fake add-haze image using a locally affine transform carried by \textbf{M} matrix. The \textbf{M} is the image information matrix including the spatial structural information of original image. In order to accurately learn the profile of objects, the paired dataset is also used to constrain the network to generate the approximate results. A paired L1 loss is added, which calculates L1 loss (red part in Fig. 2) between paired ground truth haze-free image and fake dehaze image to constraint the dehaze generator. The dehaze discriminator is used to distinguish the fake dehaze images and the real haze-free images, and it is constrained by the adversarial Loss.

In the forward constraint cycle, the network is optimized by a loss function combining the adversarial loss, the cycle consistency loss, the photorealism loss and the paired L1 loss. The whole process in Fig. 2(a) is called forward constraint cycle. There is also a backward constraint cycle presented in Fig. 2(b). For a haze-free image as input, it goes through the add-haze generator and the output image is passed into the dehaze generator to get the fake haze-free image. The add-haze discriminator is used to distinguish the fake add-haze images and the real haze images. The loss function of the backward constraint cycle is composed in the similar manner with that of the forward constraint cycle. After the above two cycle optimizations, the dehaze generator is what we ultimately want.

In our SCGAN method, there are four types of loss functions, which are the adversarial loss, the cycle consistency loss, the photorealism loss and the paired L1 loss. As our SCGAN model is trained using both unpaired dataset and paired dataset, therefore the adversarial loss is calculated for both datasets. The cycle consistency loss and the photorealism loss are measured based on the unpaired dataset, while the paired L1 loss is measured based on the paired dataset. Details of our loss functions are introduced in following five sub-sections which are corresponding to adversarial loss, cycle consistency loss based on unpaired dataset, photorealism loss based on unpaired dataset, paired L1 loss based on paired dataset, and the full loss function. These loss functions are introduced mainly for the forward constraint cycle, while those of the backward constraint cycle have similar forms.

\subsection{Adversarial Loss}
Our goal is to learn mapping functions between haze image set $X$ and haze-free image set $Y$ given training samples $X=\{{{x}_{k}}\}_{k=1}^{N}$ and $Y=\{{{y}_{k}}\}_{k=1}^{N}$. $N$ is the number of images in the dataset. In other words, given an input haze image $x$ ($x\in X$) or haze-free image $y$ ($y\in Y$), our work aims to let generators to generate output dehaze image ${y'}$ or haze image ${x'}$, respectively. The model will train two generators, the dehaze generator ${{G}_{Y}}$ for dehaze mapping $X\to Y$, and the add-haze generator ${{G}_{X}}$ for add-haze mapping $Y\to X$, and two discriminators ${{D}_{X}}$ and ${{D}_{Y}}$ that distinguish between real images from datasets and the results generated by the corresponding generators. Adversarial Loss is applied to constraint the generators and discriminators. For the mapping $X\to Y$, the objective is to produce higher quality results by solving the following optimization problem:
\begin{equation}
\begin{aligned}
{\ell_{LSGAN}}(X,Y|{G_Y},{D_Y})& = {E_{y\sim{p_{data}}(Y)}}[{({D_Y}(y) - 1)^2}]\\
& + {E_{x\sim{p_{data}}(X)}}[{D_Y}{({G_Y}(x))^2}],
\end{aligned}
\end{equation}
where ${{G}_{Y}}$ tries to minimize the objective function, by contrast, ${{D}_{Y}}$ tries to maximize it. Here, we use the adversarial loss from least square GAN (LSGAN) which replaces the calculation of the negative log likelihood with least square \cite{mao2016least}. The objective is to train a generator ${{G}_{Y}}$ to product the dehaze image $y'$ which is quite similar to images in haze-free image dataset $Y$, and also trains a discriminator ${{D}_{Y}}$ accurately distinguish $y'$ from real image $y$ in domain $Y$. The add-haze model produces the haze image deduced by the same way.

The adversarial loss is calculated for both unpaired dataset and paired dataset. The adversarial loss for unpaired dataset is ${\ell_{LSGAN}}({X_1},{Y_1}|{G_Y},{D_Y})$, where ${X}_{1}$ and ${Y}_{1}$ are the haze image set and haze-free image set in the unpaired dataset. The adversarial loss for paired dataset is ${\ell_{LSGAN}}({X_2},{Y_2}|{G_Y},{D_Y})$, where ${X}_{2}$ and ${Y}_{2}$ are the haze image set and haze-free image set in the paired dataset.

\subsection{Cycle Consistency Loss based on Unpaired Dataset}
Unpaired dataset contains natural haze images and natural outdoor haze-free images. The cycle consistency loss is calculated to constrain the model.
The input image is processed by two generators $G_Y$ and $G_X$, and the output result and the original image are compared in the cycle consistency loss to ensure individual input $x$ map to a corresponding output image $x'$. It ensures that the middle generated image of the whole process is the corresponding required transform. In the forward cycle process, it compares an image produced by dehaze generator ${{G}_{Y}}$ and add-haze generator ${{G}_{X}}$ with the original image $x$ to calculate the consistency loss. Similarly, the backward cycle will optimize the model by transforming a haze-free image to a haze image by ${{G}_{X}}$ and then get a dehaze image $y'$ by ${{G}_{Y}}$ and compare it with original haze-free image $y$. The cycle consistency loss function is defined as follows:
\begin{equation}
\begin{aligned}
{\ell _{cyc}}({X_1},{Y_1}|{G_X},{G_Y})& = {E_{x\sim{p_{data}}({X_1})}}[||{G_X}({G_Y}(x)) - x|{|_1}]\\ &+{E_{y\sim{p_{data}}({Y_1})}}[||{G_Y}({G_X}(y)) - y|{|_1}],
\end{aligned}
\end{equation}
where ${{X}_{1}}$ and ${{Y}_{1}}$ are the haze image set and haze-free image set of the unpaired dataset, respectively. The cycle consistency loss calculates the error of the generated
fake image and the original image, and transmits the loss and optimizes the generators ${{G}_{X}}$ and ${{G}_{Y}}$.

\subsection{Photorealism Loss based on Unpaired Dataset}
In the transformation by the two generators ${{G}_{Y}}$ and ${{G}_{X}}$ based on only cycle consistency constraint, some detailed textures are blurred and smoothed. In order to preserve the content consistency, the photorealism loss term is added to refine the edges, correct the color-shift and reduce the distorted effect.
Inspired by the photo style transfer \cite{luan2017deep}, Laplacian Matting is used to refine the output image of the neural network based on the original image. This photorealism regularization scheme is applied to preserve the structure of the input image and to produce photorealistic outputs. After the cycle consistency transformation, the vector form of the fake haze image denotes as $V$, multiplies with the \textbf{M} matrix which contains the related spatial structural information of original image. We minimize the following photorealism loss function:
\begin{equation}
\begin{aligned}
&{\ell_M}({X_{\rm{1}}}|{G_X},{G_Y}) = \\
&{E_{x\sim{p_{data}}}}({X_1})[V[{G_X}({G_Y}(x))]{M_x}V[{G_X}({G_Y}(x))]],
\end{aligned}
\end{equation}
where matrix $M_{x}$ contains the relative reference edge position of the input haze image $x$ which is defined as:
\begin{equation}
\begin{aligned}
&{{M}_{x}}=\underset{k|(i,j)\in {{w}_{k}}}{\mathop{\sum }}\,({{\delta }_{ij}}-\\
&\frac{1}{\text{ }\!\!|\!\!\text{ }{{w}_{k}}\text{ }\!\!|\!\!\text{ }}(1+{{({{x}_{i}}-{{\mu }_{k}})}^{T}}{{({{\sum }_{k}}+\frac{\varepsilon }{|{{w}_{k}}|}{{\text{U}}_{\text{3}}})}^{-1}}({{x}_{j}}-{{\mu }_{k}}))),
\end{aligned}
\end{equation}
where ${{x}_{i}}$ and ${{x}_{j}}$ are the colors of the input image $x$ at pixels $i$ and $j$, ${{\delta }_{ij}}$ is the Kronecker delta, ${{\mu }_{k}}$ and ${{\sum }_{k}}$ are the mean and covariance matrix of the colors in a window, ${{w}_{k}}$ is a 3$\times$3 identity matrix of the window, ${{\text{U}}_{\text{3}}}$ is a regularizing parameter, $\varepsilon$ is a regularizing parameter, and ${\rm{ }}|{{w}_{k}}|$ is the number of pixels in the window ${{w}_{k}}$. $M$ is the Matting Laplacian matrix proposed by Levin et al. \cite{Levin2008A}, which only depends on the input image.

Because the back propagation optimization needs to calculate the derivative of variables, therefore, the derivative equation of the photorealism loss is defined as follows:
\begin{equation}
\begin{aligned}
\frac{{d{\ell _M}({X_1}|{G_X},{G_Y})}}{{dV[{G_X}({G_Y}(x))]}} = 2{M_x}V[{G_X}({G_Y}(x))].
\end{aligned}
\end{equation}

The model after refined by \textbf{M} matrix manages to capture the sharp edge discontinuities and outline the profile of the objects. However, as the haze image has less edge information, this regularization is actually applied mainly for the backward constraint cycle which is constrained between haze-free images and their final fake dehaze images.

\subsection{Paired L1 Loss based on Paired Dataset}
The model trained by unpaired dataset could generate images with blue sky, however, the objects in the foreground region is hard to get proper dehaze effectiveness. To accurately learn the haze concentration in the foreground, we also add a module of using paired dataset to train the model. Each pair image contains a haze-free image and a corresponding haze image.

Pixel-wise L1-norm between the predicted and ground truth images is calculated as:
\begin{equation}
\begin{aligned}
{\ell _{L1}}({X_{\rm{2}}},{Y_{\rm{2}}}|{G_Y}) = {E_{x,y\sim{p_{data}}({X_{\rm{2}}},{Y_{\rm{2}}})}}[||{G_Y}(x) - y|{|_1}],
\end{aligned}
\end{equation}
where $X_{2}$ and $Y_{2}$ denote the haze image set and haze-free image set of the paired dataset, respectively. In the fact, L1 loss is used rather than L2 as L1 encourages less blurring and could restore highly accurate textures \cite{pathak2016context}. While this simple L1 loss encourages the generator to produce a rough outline of the predicted scene, it often fails to capture any high frequency detail. By adding an adversarial loss with the paired L1 loss will prevent this issue. The discriminator detects the true or fake images to push the generator to generate natural images.

\subsection{Full Loss Function}

The loss function based on unpaired image dataset includes three loss terms, which is expressed as follows:
\begin{equation}
\begin{aligned}
&{{\ell}_{Unpaired}}({X_1},{Y_1}|{G_X},{G_Y},{D_Y})={{\ell }_{LSGAN}}({{X}_{1}},{{Y}_{1}}|{{G}_{Y}},{{D}_{Y}})\\
&+{{\lambda }_{1}}{{\ell }_{cyc}}({{X}_{1}},{{Y}_{1}}|{{G}_{X}},{{G}_{Y}})+{{\lambda }_{2}}{{\ell }_{M}}({{X}_{1}|{{G}_{X}},{{G}_{Y}}}).
\end{aligned}
\end{equation}
We linearly combine the three items according to scales  ${{\lambda }_{1}}$ and ${{\lambda }_{2}}$ , and try to make every item plays a role to finally work out the optimal results.

The loss function based on paired dataset is:
\begin{equation}
\begin{aligned}
{\ell _{Paired}}({X_{\rm{2}}},{Y_{\rm{2}}}|{G_Y},{D_Y})& = {\lambda _3}{\ell _{LSGAN}}({X_{\rm{2}}},{Y_{\rm{2}}}|{G_Y},{D_Y})\\
&+ {\lambda _4}{\ell _{L1}}({X_{\rm{2}}},{Y_{\rm{2}}}|{G_Y}),
\end{aligned}
\end{equation}
where ${{\lambda }_{3}}$ and ${{\lambda }_{4}}$ are two constant coefficients. The optimization based on paired dataset accurately learns haze features. The concentration of haze is controlled by the depth of image. The degree of dehazing in foreground can also be controlled preciously and effectively to avoid saturation problem in the foreground region. In conclusion, by combining paired images training, the authenticity of the whole image is improved.

By combining two datasets and their appropriate loss constraints, we can obtain the proper solution. The total loss function is:
\begin{equation}
\begin{aligned}
&{\ell _{Total}}({X_1},{Y_1},{X_2},{Y_2}|{G_X},{G_Y},{D_Y})=\\
&{\ell_{Unpaired}}({X_1},{Y_1}|{G_X},{G_Y},{D_Y})
+ {\ell _{Paired}}({X_2},{Y_2}|{G_Y},{D_Y}),
\end{aligned}
\end{equation}
where ${{\ell }_{Total}}({X_1},{Y_1},{X_2},{Y_2}|{G_X},{G_Y},{D_Y})$ is the total loss function of the forward constraint cycle. The total loss function of the backward constraint cycle is ${{\ell }_{Total}}({X_1},{Y_1},{X_2},{Y_2}|{G_X},{G_Y},{D_X})$, which is defined in the same manner with that of the forward constraint cycle. So the generator which we finally require is the dehaze generator ${{G}_{Y}}$.  The objective is solved by:
\begin{equation}
\begin{aligned}
{{G}_{Y}}=\arg \underset{{{G}_{Y}},{{G}_{X}}}{\mathop{\min }}\,\underset{D{}_{Y},{{D}_{X}}}{\mathop{\max }}\,\ell ({{G}_{Y}},{{G}_{X}},{{D}_{Y}},{{D}_{X}}),
\end{aligned}
\end{equation}
where $\ell ({{G}_{Y}},{{G}_{X}},{{D}_{Y}},{{D}_{X}})$ is the full loss function combining the total loss of the forward constraint cycle and that of the backward constraint cycle.

\section{Implementation Details}
\subsection{Dataset Collection}
\label{subsec:data}
10000 haze images were downloaded from the Google images using the tags {\em haze}, {\em smog}, {\em fog}. 10000 haze-free images were also collected by searching using tags {\em clear day city}, {\em landscape}, {\em blue sky}, {\em sunny weather}, {\em sunny city}, {\em street view. etc}. These 10000 haze images and 10000 haze-free images form the unpaired dataset.

The paired image dataset is created by adding haze on the haze-free images which from the Make3D datasets \cite{saxena20083,saxena2009make3d}. These haze-free images have their corresponding depth information. We use the atmospheric physical model to calculate the transmission value based on depth information. The image degradation model is widely used in computer vision and image processing, which is described as:
\begin{equation}
I(p)=J(p)t(p)+A(1-t(p)),
\end{equation}
where $I$ is the observed intensity with haze, $J$ is the scene radiance without haze, $t$ is the medium transmission describing the portion of the light that is not scattered and reaches the camera, and $A$ is the global atmospheric light. The transmission $t$ is determined by the distance which the light travels from the object to the camera. The longer the distance, the greater the attenuated \cite{narasimhan2002vision}. Therefore, the transmission can be expressed as:
\begin{equation}
t(p)={{e}^{-\beta d(p)}},
\end{equation}
where $d$ is the depth of the scene, and $\beta$ is the scattering coefficient which is set as constant as defined in \cite{zhu2017haze,Yoon2015Wavelength}. When we add haze on the haze-free images, the atmospheric value $A$ is not set as 1, because the general haze is not pure white, it has slightly grey, here we set it as 0.85. The depth information of each image is captured by an infrared sensor, which is coarse. Therefore, instead of directly using the original depth information, we use a Laplace matting processing \cite{Levin2008A} to filter the depth map, and obtain the refined depth map that the boundary approximates the original image. The principle formula is Equation 4. Here we use the closed solution to solve the function. The details of the algorithm refer to \cite{Levin2008A}.

\subsection{Training Details}
In this experiment, the entire network is trained on a NVIDIA P100 GPU using the Torch framework \cite{collobert2011torch7}, using Adam algorithm \cite{Kingma2014Adam} with a learning rate of 2$\times10^{-5}$ and the momentum of 0.9 for the first 200 epochs and a linearly decaying rate that goes to zero over the next 200 epochs. The weights in the model are initialized by mean 0 and standard deviation 0.02 of Gaussian distribution. For fair comparison, in the following experiments, if there are no special instructions, the model is trained with 1000$\times$2 unpaired images and 340$\times$2 paired images of totally 2680 images. The images are scaled to 256$\times$256 pixels. The training requires about 2200MB GPU memory, and it costs 38 hours for training in 400 iterations. While training using 20000 unpaired images, it takes more than one week. In the optimization, our algorithm consists of jointly minimizing and maximizing conflicting objectives, it is not stable in the training procedure. Therefore, we alternately train the model by using the two datasets, which is to train the model using unpaired images in one iteration and then train it using paired images in another iteration.
\begin{figure*}[t]
\centering
\subfigure[Original]{\includegraphics[width=0.13\linewidth]{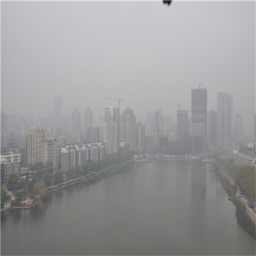}}
\subfigure[D:C+A]{\includegraphics[width=0.13\linewidth]{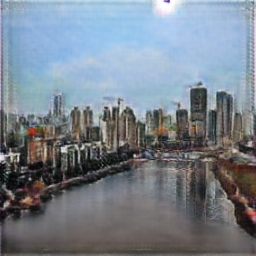}}
\subfigure[D:C+A+M]{\includegraphics[width=0.13\linewidth]{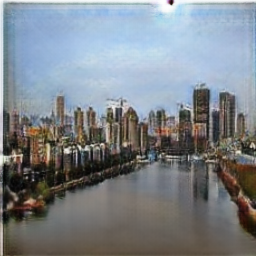}}
\subfigure[D:A+Paired L1]{\includegraphics[width=0.13\linewidth]{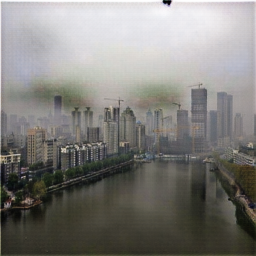}}
\subfigure[D:C+A+Paired L1]{\includegraphics[width=0.13\linewidth]{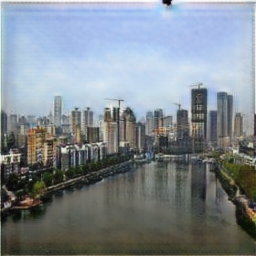}}
\subfigure[D:C+A+Paired L1+M]{\includegraphics[width=0.13\linewidth]{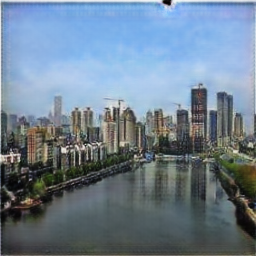}}\\
\includegraphics[width=0.13\linewidth]{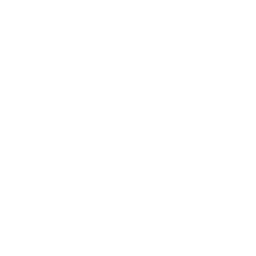}
\subfigure[A:C+A]{\includegraphics[width=0.13\linewidth]{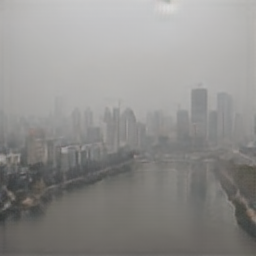}}
\subfigure[A:C+A+M]{\includegraphics[width=0.13\linewidth]{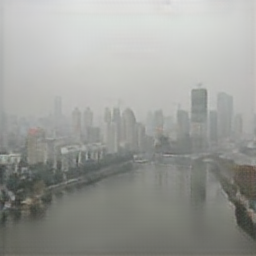}}
\subfigure[A:A+Paired L1]{\includegraphics[width=0.13\linewidth]{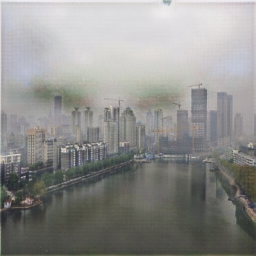}}
\subfigure[A:C+A+Paired L1]{\includegraphics[width=0.13\linewidth]{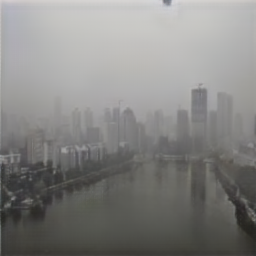}}
\subfigure[A:C+A+Paired L1+M]{\includegraphics[width=0.13\linewidth]{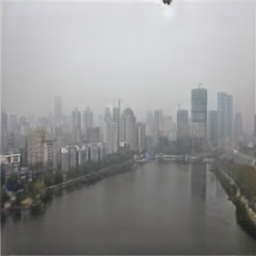}}
\caption{Effect of image dehaze by using different loss function to train our model. (a)Original image. (b-f) The dehaze results trained with different loss constraint items (D: Dehaze result. M: Photorealism loss). (g-k) Results of add-haze effect based on corresponding fake dehaze image in (b)-(f) (A: Add-haze result). C+A model is trained using adversarial loss and cycle consistency loss of unpaired dataset. A+Paired L1 model is trained using adversarial loss and paired L1 loss based on paired images only. C+A+Paired L1 is trained using adversarial loss, cycle consistency loss based on unpaired dataset, and paired L1 loss based on paired dataset. C+A+Paired L1 is trained using our four kinds of loss functions.}
\label{fig:onecol}
\end{figure*}
\subsection{Network Architecture}
\label{sbsec:c}
The generators use a simple encoder-decoder structure and have two residual blocks in the middle of two coders as shown in Fig. 2. The first convolution and the last convolution layers use 7$\times$7 convolution kernel and stride 1, other convolution layers use 3$\times$3 convolution kernel and stride 2, and the corresponding deconvolution layer has stride of 1/2. Each convolution layer attaches a batch-normalization layer (Norm) and a ReLU layer in the back. Here the batch is set as 1, and instance normalization \cite{Ulyanov2016Instance} is used.

For the discriminator architecture, we use 70$\times$70 PatchGAN \cite{johnson2016perceptual}, \cite{isola2016image} which discriminates local patches rather than a whole image. In this way, the results are more natural. All the convolution kernels are set as $3$, and the last two convolution layers set stride as 1, others are 2. Leaky Relu layer sets slop as 0.2.

\section{Experiments}
\subsection{Effectiveness of Loss Function}

To deeply understand the effect of each part of the objective loss function, the dehaze results generated by the dehaze generator optimized by using different parts of loss function are shown in Fig. 3. Our cycle structure trained using the cycle consistency loss and adversarial loss based on unpaired dataset is called C+A model. The dehaze image produced by C+A model by using unpaired data can get a clear image. However, the image is blurred with artifacts which are produced by fake edges (see Fig. 3(b)). After adding photorealism loss in Fig. 3(c), the dehaze image quality is improved. However, there are still artifacts created by fake edges. The reason is that the difference between the final output image and the original haze image is very large, and it makes the photorealism loss to be hard to converge. Thus the boundaries are not refined well. The dehaze result in Fig. 3(d) is generated based on the model trained using adversarial loss and paired L1 loss based on paired images only (called A+Paired L1). There are no fake edges created in the dehaze result in Fig. 3(d). However, the dehaze image still has a haze sky, which makes the haze removal effect not clear comparing to the original haze image. In the dehaze result generated by the model trained using adversarial loss, cycle consistency loss of unpaired data and paired L1 loss (Fig. 3(e)), there is created clear profile, and the sky is very clear. However, the textures of objects are smoothed with less details. After adding photorealism loss in the objective function (Fig. 3(f)), the details of the image are changed to much clear with less fake edges. The color is also much more natural.

The effort of add-haze also could reflect the effect of different loss functions. In the third row of Fig. 3, the final add-haze image (Fig. 3(k)) is the closest to the original image (Fig. 3(a)). The add-haze image generated by using the model without photorealism loss in Fig. 3(j) suffers the phenomenon of loss of texture information. To quantitatively analyze the effect of loss functions, Peak Signal to Noise Ratio (PSNR) is used to measure the similarity of the add-haze images and the original haze images. The add-haze images are generated by different models constrained by different loss terms. 300 real-world haze images collected from the Internet are used for this analysis. The boxplot of PSNR shows that the overall score of our final loss is superior to scores of other loss terms in Fig. 4. The red line in the middle of the box in the rightmost box is higher than others. With a few exceptions, the maximum value (the short and black line on the top of the box object), the minimum (the short and black line on the bottom of the box object), one fifth (the blue line of the upper edge of the box), four fifth (the blue line of the under edge of the box) of the rightmost box are also higher than the other box. These represent that the add-haze results by our final loss function are the most closest to the original images comparing to results generated by models trained using other loss combinations. Additionally, by comparing the PSNR values of model Cyc+L and model Cyc+L+M in Fig. 4, we also observe that by adding the photorealism loss in our model, the PSNR values are distinctly improved.
\begin{figure}[t!]
\begin{center}
   \includegraphics[width=0.8\linewidth]{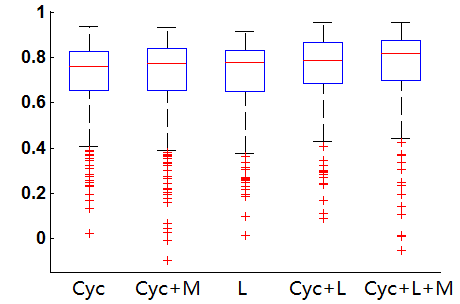}
   \end{center}
   \caption{The distribution of PSNR values which calculate the similarity of the original haze images with the add-haze results. The add-haze result is generated by passing the original image through a dehaze generator and then an add-haze generator using different loss terms (Cyc: C+A loss, L: A+Paired L1 loss, M: Photorealism loss). The red line in the middle of the box is the median value of the data. The short and black line on the top of the box object is the maximum value, and the short black bottom line of the box is the minimum value (excluded outliers in red points). The blue lines of the upper edge and under edge are the one-fifth and four-fifth lines.}
\end{figure}

Here we also analyze the effect of photorealism loss in the backward constraint cycle process, which means that a haze-free image is first added haze and then is dehazed. The add-haze process plays an important role in the whole model. If the add-haze has bad performance, the cycle consistency constraint is hard to converge. This will also affect the dehaze performance and it is difficult to refine the profile of the object.
\begin{figure*}[t!]
\begin{center}
   \includegraphics[width=0.14\linewidth]{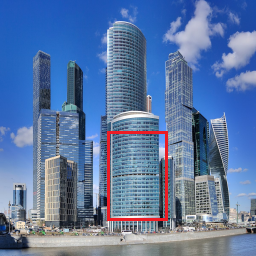}
   \includegraphics[width=0.14\linewidth]{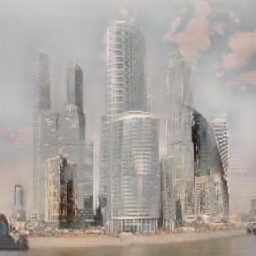}
   \includegraphics[width=0.14\linewidth]{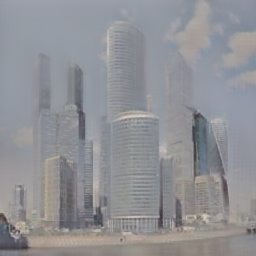}
   \includegraphics[width=0.14\linewidth]{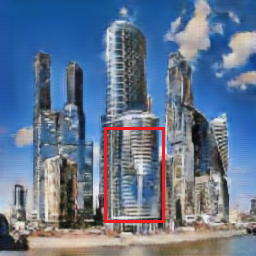}
   \includegraphics[width=0.14\linewidth]{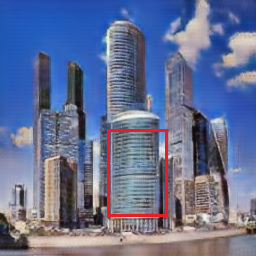}\\
   \includegraphics[width=0.14\linewidth]{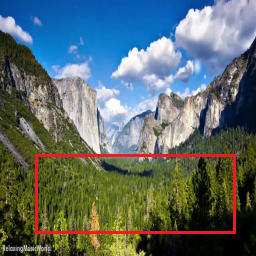}
   \includegraphics[width=0.14\linewidth]{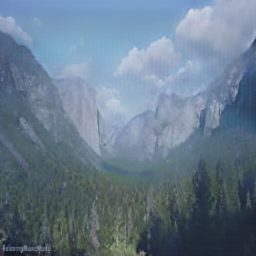}
   \includegraphics[width=0.14\linewidth]{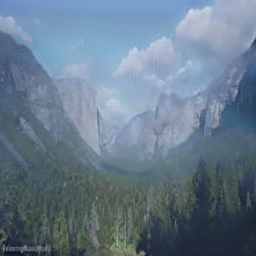}
   \includegraphics[width=0.14\linewidth]{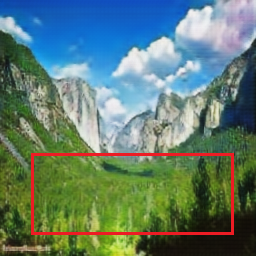}
   \includegraphics[width=0.14\linewidth]{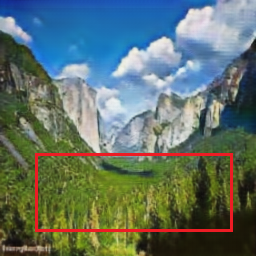}\\
   \subfigure[]{\includegraphics[width=0.14\linewidth]{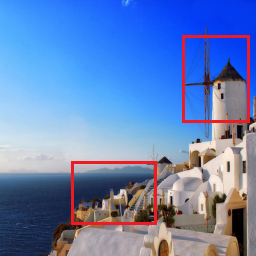}}
   \subfigure[]{\includegraphics[width=0.14\linewidth]{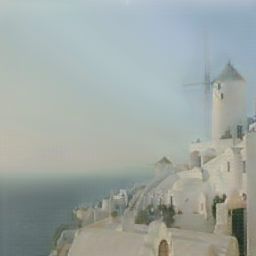}}
   \subfigure[]{\includegraphics[width=0.14\linewidth]{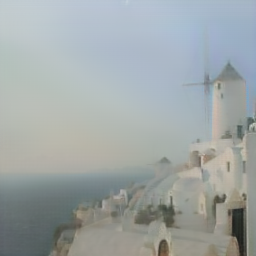}}
   \subfigure[]{\includegraphics[width=0.14\linewidth]{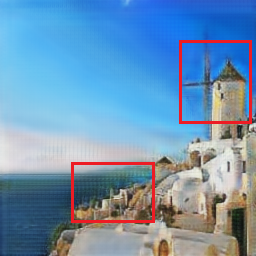}}
   \subfigure[]{\includegraphics[width=0.14\linewidth]{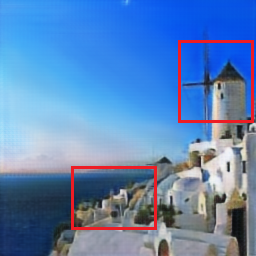}}
   \end{center}
   \caption{Backward constraint cycle process (The haze-free image is first added haze and then is dehazed). In red boxes, the effect of the distinction is obvious. (a)Original image. (b)Add-haze effect without photorealism loss constraint. (c)Add-haze effect by using our SCGAN. (d)Dehazing (b) by using dehaze generator without photorealism loss constraint. (e)Dehazing (c) by using our SCGAN dehaze generator.}
\end{figure*}
Some example results in the backward constraint cycle process generated by using our models with and without photorealism loss are shown in Fig. 5. By comparing results in Fig. 5(d) and (e), it shows that our SCGAN method with the photorealism loss performs better than that without the photorealism loss. The photorealism loss could avoid artificial edges especially the textures in the red box. It generates images with clear edges and the color of object is consistent. Therefore, the photorealism loss could successfully avoid the artificial edge generation. The authenticity of the image is greatly enhanced.
\subsection{Parameter Selection}
\begin{figure*}[!]
\begin{center}
   \subfigure[]{\includegraphics[width=0.13\linewidth]{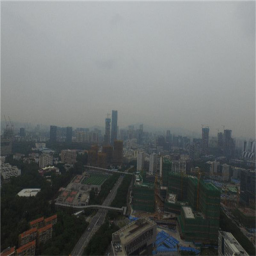}}
   \subfigure[${{\lambda }_{2}}$ = 0]{\includegraphics[width=0.13\linewidth]{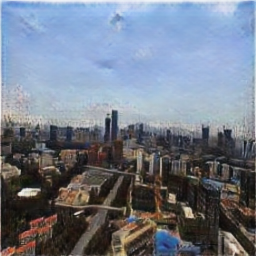}}
   \subfigure[${{\lambda }_{2}}$ = 1]{\includegraphics[width=0.13\linewidth]{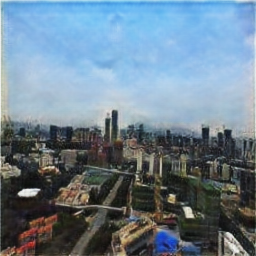}}
   \subfigure[${{\lambda }_{2}}$ = 2]{\includegraphics[width=0.13\linewidth]{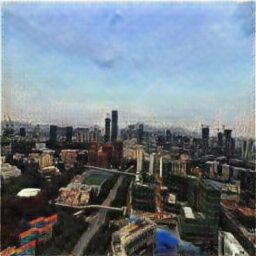}}
   \subfigure[${{\lambda }_{2}}$ = 3]{\includegraphics[width=0.13\linewidth]{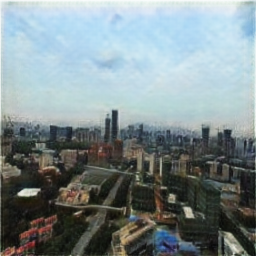}}
   \subfigure[${{\lambda }_{2}}$ = 5]{\includegraphics[width=0.13\linewidth]{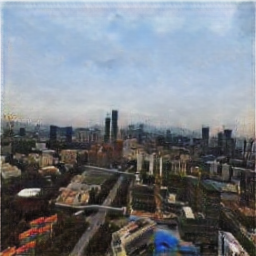}}
   \subfigure[${{\lambda }_{2}}$ = 10]{\includegraphics[width=0.13\linewidth]{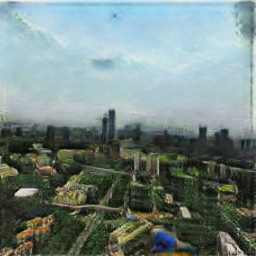}}
   \end{center}
   \caption{The effect of different ${{\lambda }_{2}}$ values. (a)Original image. (b-g)Dehaze effects by setting  ${{\lambda }_{2}}$ = 0, 1, 2, 3, 5, 10,respectively.}
\end{figure*}
\begin{table}
\caption{OBJECTIVE ASSESSMENT OF PARAMETER SETTING. AVERAGE PSNR AND SSIM VALUES OF 300 PARIES OF RESULTING ADD-HAZE IMAGES AND ORIGINAL HAZE IMAGES.}
\begin{center}
\begin{tabular}{|l|c|c|c|c|c|c|}
\hline\hline
 ${{\lambda }_{2}}$ & 0 & 1 & 2 & 3 & 5 & 10\\
\hline
PSNR & 22.31& 27.17 & 26.77 & \textbf{29.23} & 24.67 & 24.59\\
SSIM & 0.85 & \textbf{0.91} & 0.90 & 0.88 & 0.88 & 0.85\\
\hline
\end{tabular}
\end{center}
\end{table}
Our model contains four important parameters. ${{\lambda }_{1}}$ controls the cycle consistency loss value. It is set as 10, which is the same as the CycleGAN model. ${{\lambda }_{2}}$ controls the photorealism loss value. Based on experimental observation, we found that we could get the best effect by setting it as 1 to 3. Small ${{\lambda }_{2}}$ will not learn the realism photo features and the edges of objects have artificial bending and shadow (see Fig. 6). Large value of it will cause color-shift. Fig. 6(g) shows that the edge is not distinct and the color is strange while using large ${{\lambda }_{2}}$. Table \uppercase\expandafter{\romannumeral1} shows the average PSNR and the structure similar degree (SSIM) \cite{Wang2004Image} values of 300 paries of resulting add-haze images and original haze images. These add-haze images are generated by using different parameter setting generators. From it, we also find that setting ${{\lambda }_{2}}$ in 1 to 3 will get the best performance. When the parameter across over the optimal range, the quality of generated images reduces. ${{\lambda }_{3}}$ and ${{\lambda }_{4}}$ assembly affect the paired loss values. They are set in the same way as in Pix2pixGAN model \cite{isola2016image}. ${{\lambda }_{3}}$ is set as 9.9 in our experiments, and ${{\lambda }_{4}}$ is set as 0.1.
\subsection{The Effect of Training Dataset Size}
\begin{figure*}[!]
\begin{center}
   \subfigure[]{\includegraphics[width=0.15\linewidth]{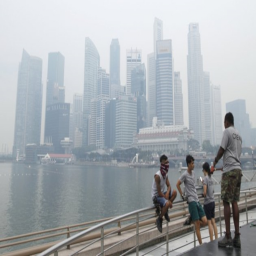}}
   \subfigure[500]{\includegraphics[width=0.15\linewidth]{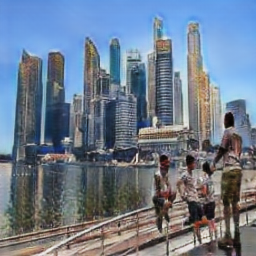}}
   \subfigure[1000]{\includegraphics[width=0.15\linewidth]{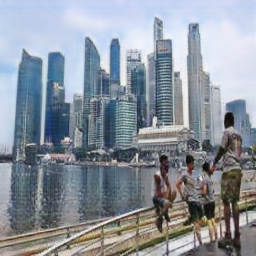}}
   \subfigure[2000]{\includegraphics[width=0.15\linewidth]{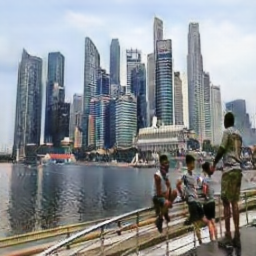}}
   \subfigure[5000]{\includegraphics[width=0.15\linewidth]{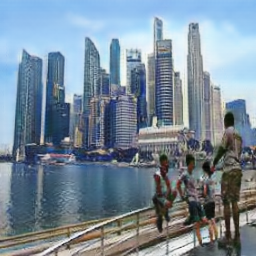}}
   \subfigure[10000]{\includegraphics[width=0.15\linewidth]{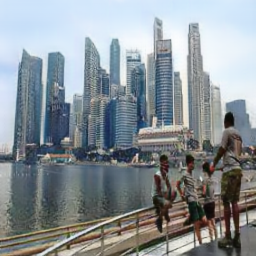}}
   \end{center}
   \caption{Effect of training dataset size. (a)Original image. (b-f)Dehaze results generated by models trained using 500, 1000, 2000, 5000, 10000 pairs of unpaired images datasets, respectively.}
\end{figure*}
\begin{table}
\caption{OBJECTIVE ASSESSMENT OF TRAINING UNPAIRED DATASET SIZE. THE NUMBER OF RESNET IS CORRESPONDING TO THE NUMBER OF RESIDUAL BLOCKS. PSNR AND SSIM VALUES ARE THE AVERAGE PSNR AND SSIM VALUES OF 300 PARIES OF RESULTING ADD-HAZE IMAGES AND ORIGINAL HAZE IMAGES.}
\begin{center}
\begin{tabular}{|l|c|c|c|c|c|}
\hline\hline
Size & 500 & 1000 & 2000 & 5000 & 10000\\
\hline
ResNet & 0 & 2 & 3 & 6 & 9\\
Cost(day) & 0.92 & 1.58 & 2.43 & 5.80 & 9.0\\
PSNR & 24.48& 25.75 & \textbf{27.29} & 24.03 & 21.55\\
SSIM & 0.77 & 0.85 & 0.81 & 0.85 & \textbf{0.86}\\
\hline
\end{tabular}
\end{center}
\end{table}
Here, we will analyze how the unpaired training dataset size affects the dehaze effect. When training the same model, it is not true that better effect is achieved by using more training data. Because the large number of data needs more parameters to fit them. We use a small network which contains 2 residual blocks to train 500 and 1000 pairs of unpaired images, and use 3 blocks to train 2000 pairs of unpaired images. 6 residual blocks are used for 5000 pairs of unpaired images and 9 residual blocks are used to fit 10000 pairs of unpaired datasets. The residual blocks are in the middle of two coders of generators. Figure 7 shows that 500 pairs unpaired dataset is too small to train a stable model. 1000 pairs dataset is not much different from 2000 or 5000 pairs. But we can see that the model trained using 10000 pairs dataset is more stable than that using 5000 pairs dataset on the ability of dehazing. There are fake edges and blur regions in the result of using 5000 pairs dataset. The 5000 dataset may not be enough to fit a network of 6 blocks. The number of residual blocks and time consumption of different datasets are summarized in Table \uppercase\expandafter{\romannumeral2}. The average PSNR and SSIM values of the 300 real-word haze images are also summarized in Table \uppercase\expandafter{\romannumeral2}. Because we change the architecture of the network to adapt for the different size of datasets, the PSNR and SSIM values do not have a consistent changing tendency along increasing the training dataset size. However, the time cost of training is dramatically increased along increasing the training image dataset size. The testing results generated by the model trained using 2000 pairs of unpaired images have the highest PSNR, those generated by the model trained using 10000 pairs of unpaired images have the highest SSIM. The testing results generated by the model trained using 1000 pairs of unpaired images have both the second highest PSNR and SSIM values, and this model needs much less training time. Therefore, in this paper we only use 1000 pairs unpaired dataset with 2 residual blocks to do the experiments. The trained model has good effect with less cost time. It is good for us to do experiments and tune parameters on the small dataset.

\subsection{Comparison with State-of-The-Art Methods}
Our method is compared to seven popular related image dehaze methods and two GAN methods. The seven related image dehaze methods are all using the atmospheric physical model. He {\em et al.} \cite{he2011single} and Berman {\em et al.} \cite{berman2016non} use different mapping calculations to replace transmission map. Cai {\em et al.} \cite{cai2016dehazenet} and Ren {\em et al.} \cite{ren2016single} use deep learning methods to extract the transmission map. Zhu {\em et al.} \cite{zhu2017haze} combines the dark channel prior and luminance prior based on the segmentation of sky region and non-sky region to conduct the haze removal. In contrast, Li {\em et al.} \cite{li2017aod} and Zhang {\em et al.} \cite{zhang2018densely} jointly learn the transmission map, atmospheric light and dehazing by using GAN model. Figure 8 shows the visual comparison of different algorithms on a few real-world examples without post-processing. Pix2pixGAN is trained by our 340 paired dataset and CycleGAN is trained by unpaired dataset with 1000 haze images and 1000 haze-free images.
\begin{figure*}[htbp]
\flushleft
Original image \ \ \ \ \
   \includegraphics[width=0.115\linewidth]{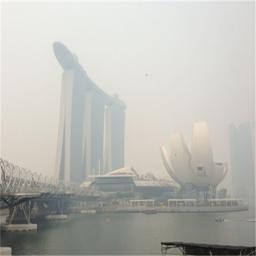}
   \includegraphics[width=0.115\linewidth]{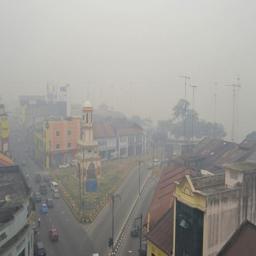}
   \includegraphics[width=0.115\linewidth]{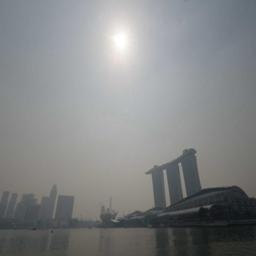}
   \includegraphics[width=0.115\linewidth]{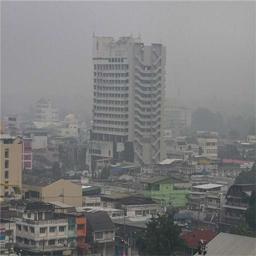}
   \includegraphics[width=0.115\linewidth]{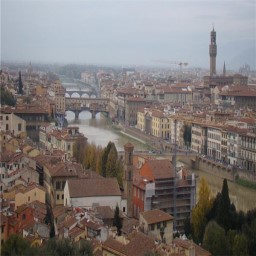}
   \includegraphics[width=0.115\linewidth]{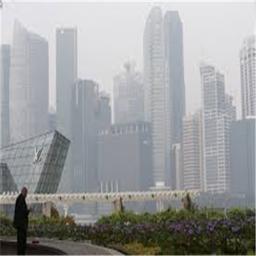}
   \includegraphics[width=0.115\linewidth]{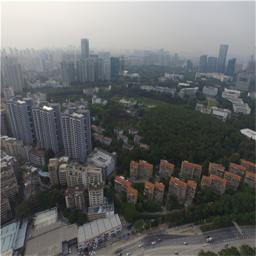}
   \\
   He {\em et al.} \cite{he2011single}\ \ \ \ \ \ \ \ \
   \includegraphics[width=0.115\linewidth]{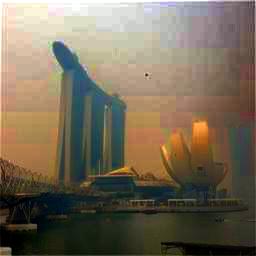}
   \includegraphics[width=0.115\linewidth]{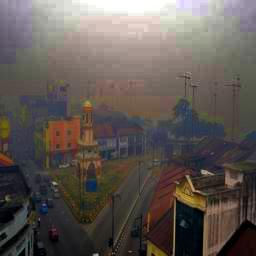}
   \includegraphics[width=0.115\linewidth]{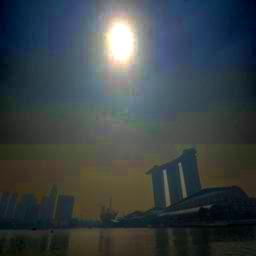}
   \includegraphics[width=0.115\linewidth]{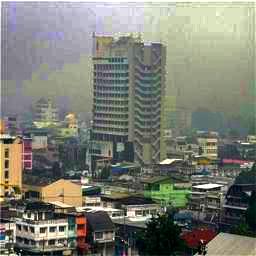}
   \includegraphics[width=0.115\linewidth]{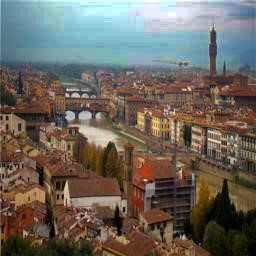}
   \includegraphics[width=0.115\linewidth]{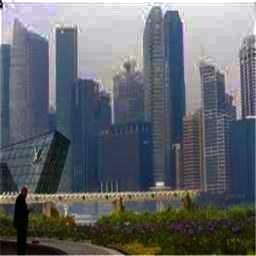}
   \includegraphics[width=0.115\linewidth]{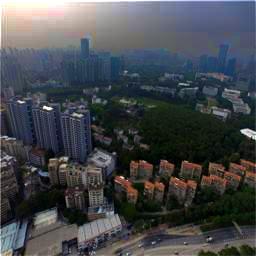}
   \\
   Berman {\em et al.} \cite{berman2016non}
   \includegraphics[width=0.115\linewidth]{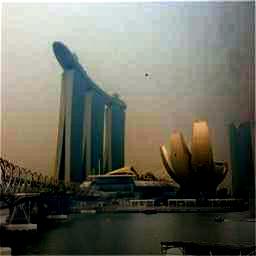}
   \includegraphics[width=0.115\linewidth]{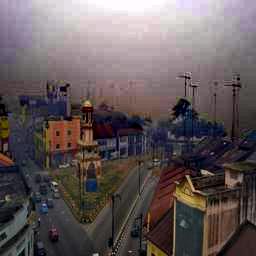}
   \includegraphics[width=0.115\linewidth]{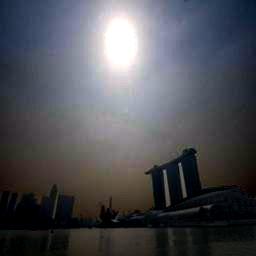}
   \includegraphics[width=0.115\linewidth]{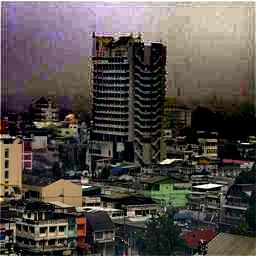}
   \includegraphics[width=0.115\linewidth]{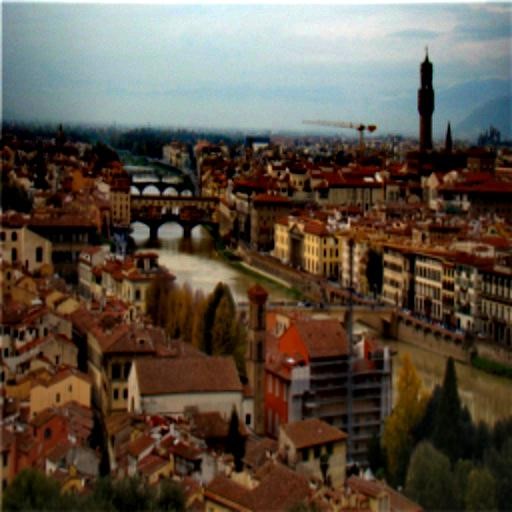}
   \includegraphics[width=0.115\linewidth]{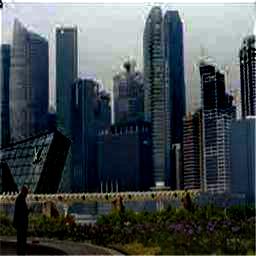}
   \includegraphics[width=0.115\linewidth]{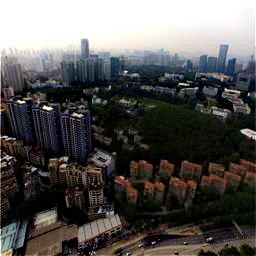}
    \\
    Cai {\em et al.} \cite{cai2016dehazenet}\ \ \ \ \ \
   \includegraphics[width=0.115\linewidth]{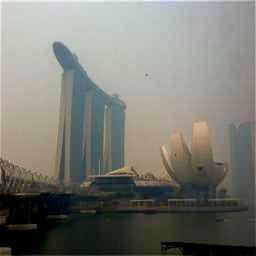}
   \includegraphics[width=0.115\linewidth]{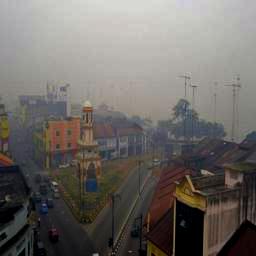}
   \includegraphics[width=0.115\linewidth]{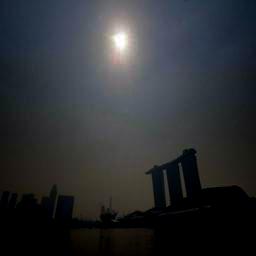}
   \includegraphics[width=0.115\linewidth]{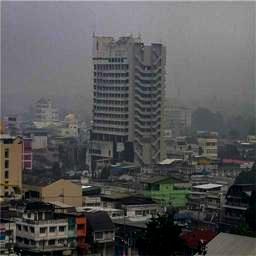}
   \includegraphics[width=0.115\linewidth]{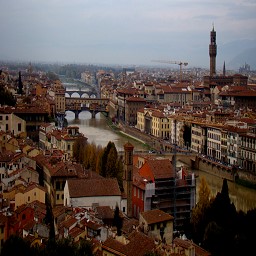}
   \includegraphics[width=0.115\linewidth]{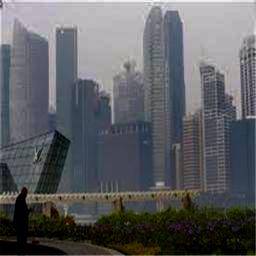}
   \includegraphics[width=0.115\linewidth]{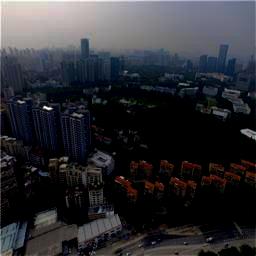}
   \\
   Ren {\em et al.} \cite{ren2016single}\ \ \ \ \ \ \
   \includegraphics[width=0.115\linewidth]{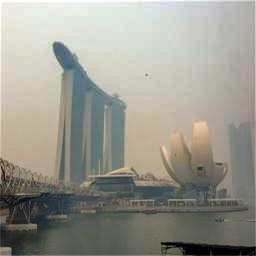}
   \includegraphics[width=0.115\linewidth]{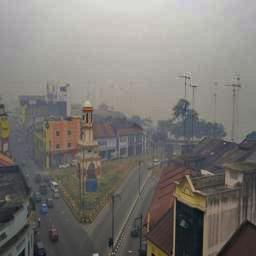}
   \includegraphics[width=0.115\linewidth]{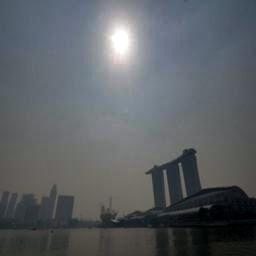}
   \includegraphics[width=0.115\linewidth]{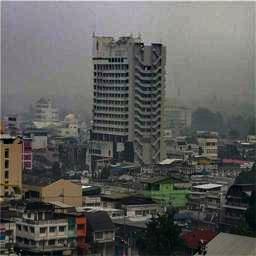}
   \includegraphics[width=0.115\linewidth]{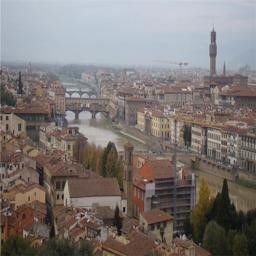}
   \includegraphics[width=0.115\linewidth]{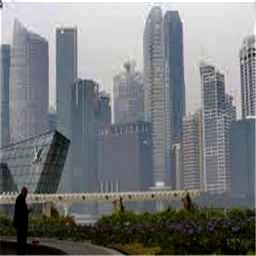}
   \includegraphics[width=0.115\linewidth]{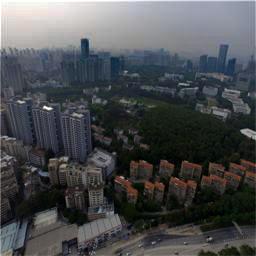}
   \\
   Zhu {\em et al.} \cite{zhu2017haze}\ \ \ \ \ \
   \includegraphics[width=0.115\linewidth]{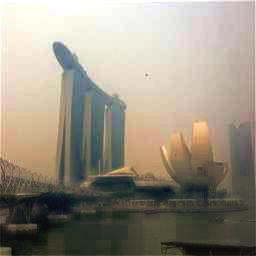}
   \includegraphics[width=0.115\linewidth]{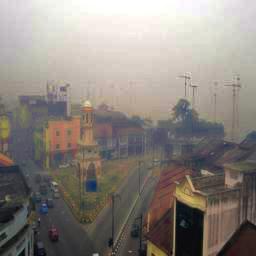}
   \includegraphics[width=0.115\linewidth]{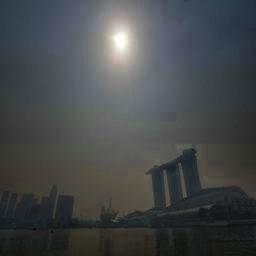}
   \includegraphics[width=0.115\linewidth]{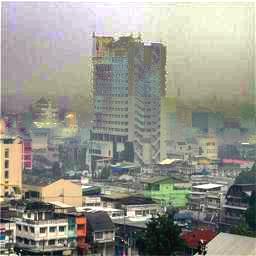}
   \includegraphics[width=0.115\linewidth]{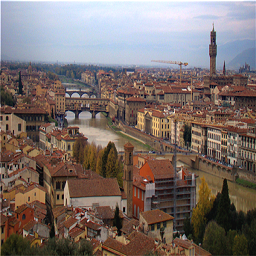}
   \includegraphics[width=0.115\linewidth]{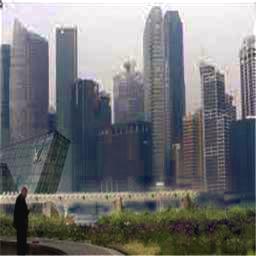}
   \includegraphics[width=0.115\linewidth]{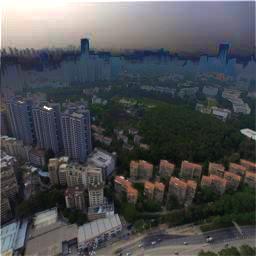}
   \\
   Li {\em et al.} \cite{li2017aod}\ \ \ \ \ \ \ \ \
   \includegraphics[width=0.115\linewidth]{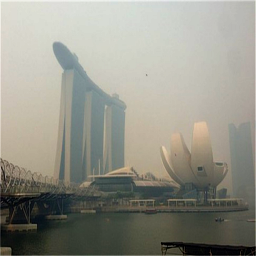}
   \includegraphics[width=0.115\linewidth]{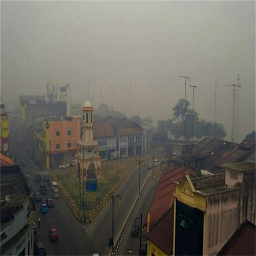}
   \includegraphics[width=0.115\linewidth]{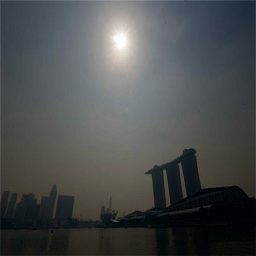}
   \includegraphics[width=0.115\linewidth]{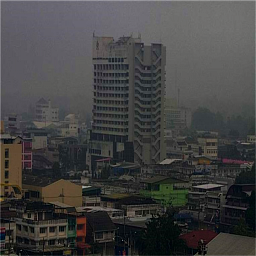}
   \includegraphics[width=0.115\linewidth]{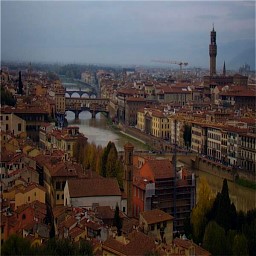}
   \includegraphics[width=0.115\linewidth]{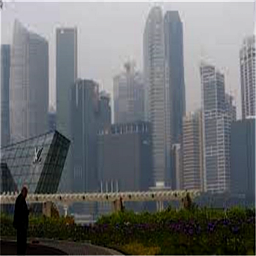}
   \includegraphics[width=0.115\linewidth]{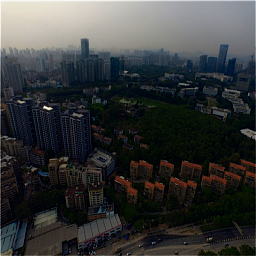}
   \\
   Zhang {\em et al.} \cite{zhang2018densely}\ \ \
   \includegraphics[width=0.115\linewidth]{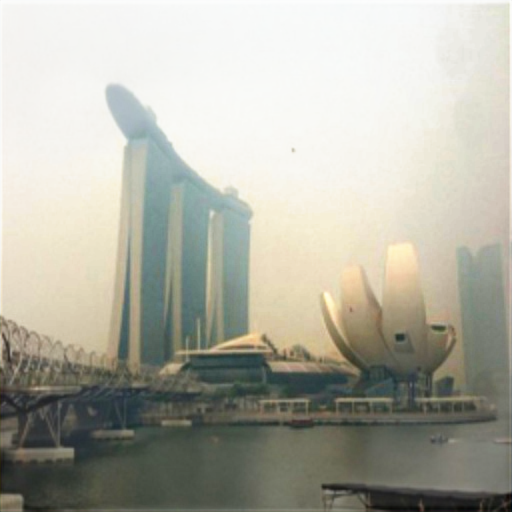}
   \includegraphics[width=0.115\linewidth]{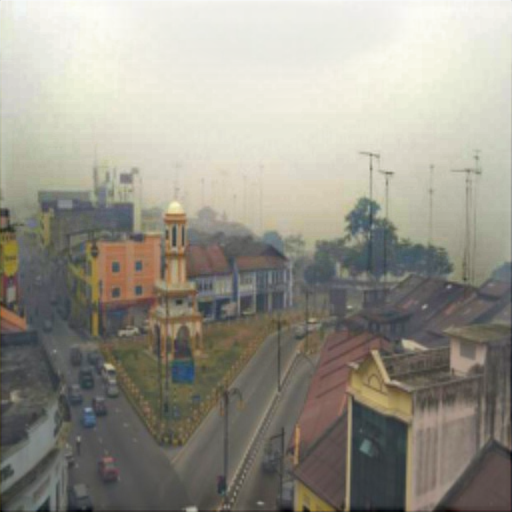}
   \includegraphics[width=0.115\linewidth]{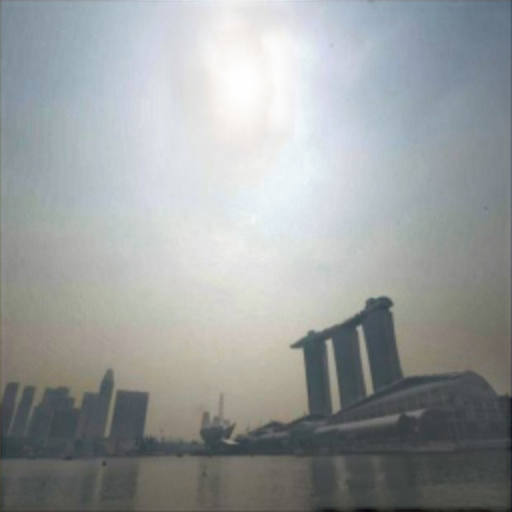}
   \includegraphics[width=0.115\linewidth]{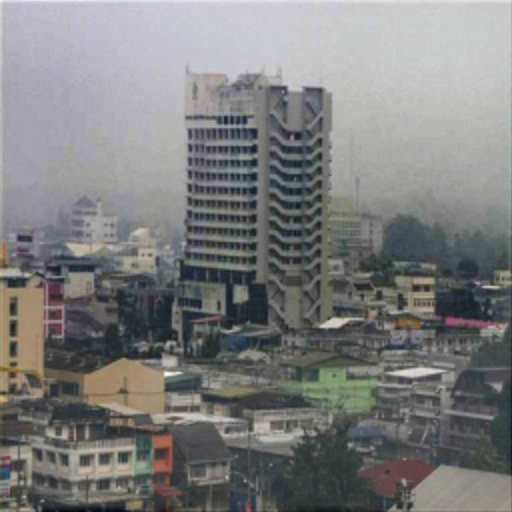}
   \includegraphics[width=0.115\linewidth]{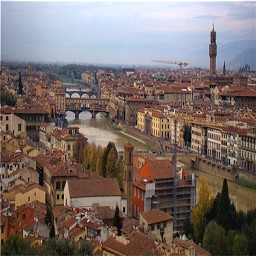}
   \includegraphics[width=0.115\linewidth]{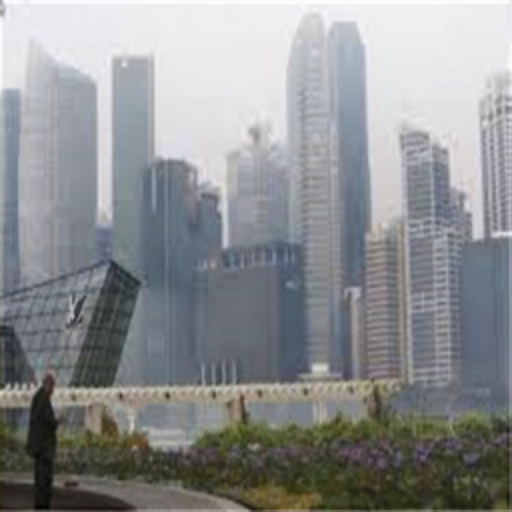}
   \includegraphics[width=0.115\linewidth]{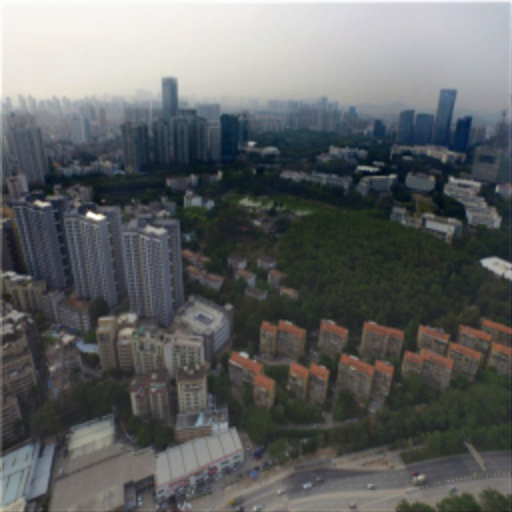}
    \\
   Pix2pixGAN \cite{isola2016image}\ \
   \includegraphics[width=0.115\linewidth]{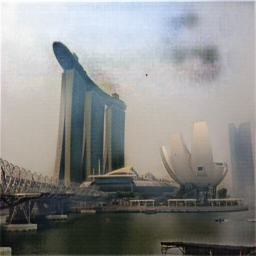}
   \includegraphics[width=0.115\linewidth]{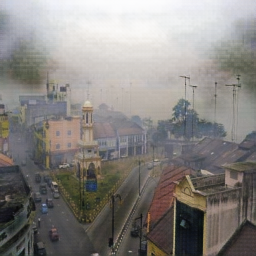}
   \includegraphics[width=0.115\linewidth]{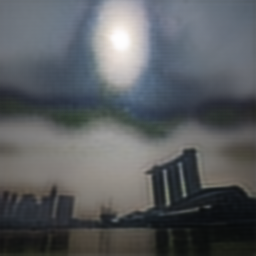}
   \includegraphics[width=0.115\linewidth]{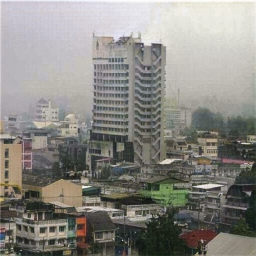}
   \includegraphics[width=0.115\linewidth]{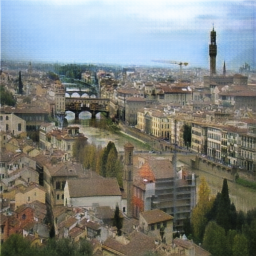}
   \includegraphics[width=0.115\linewidth]{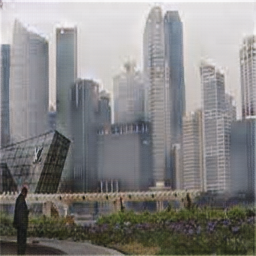}
   \includegraphics[width=0.115\linewidth]{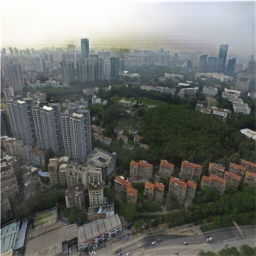}
   \\
   CycleGAN \cite{zhu2017unpaired} \ \ \
   \includegraphics[width=0.115\linewidth]{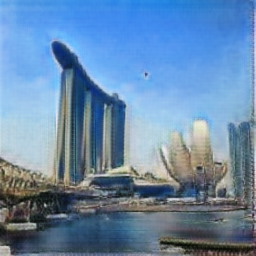}
   \includegraphics[width=0.115\linewidth]{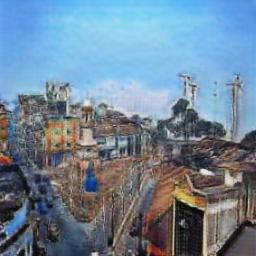}
   \includegraphics[width=0.115\linewidth]{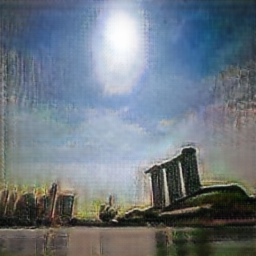}
   \includegraphics[width=0.115\linewidth]{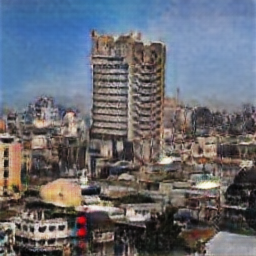}
   \includegraphics[width=0.115\linewidth]{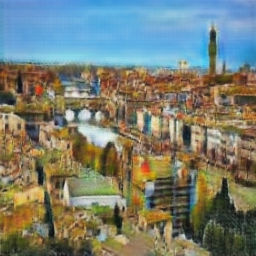}
   \includegraphics[width=0.115\linewidth]{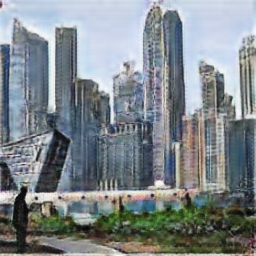}
   \includegraphics[width=0.115\linewidth]{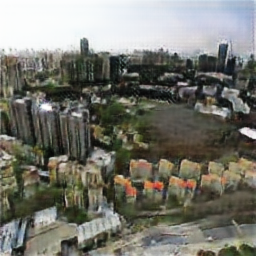}
   SCGAN \ \ \ \ \ \ \ \ \ \ \ \
   \includegraphics[width=0.115\linewidth]{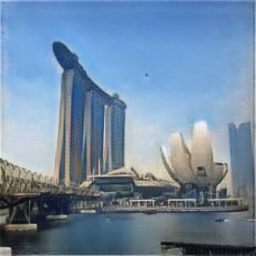}
   \includegraphics[width=0.115\linewidth]{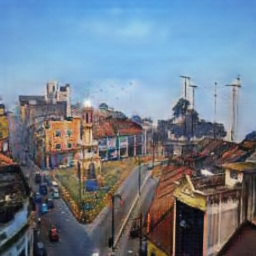}
   \includegraphics[width=0.115\linewidth]{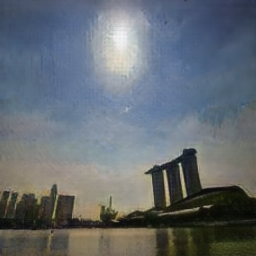}
   \includegraphics[width=0.115\linewidth]{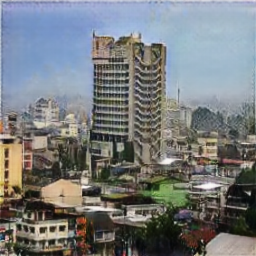}
   \includegraphics[width=0.115\linewidth]{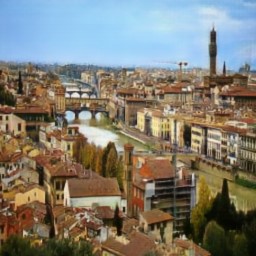}
   \includegraphics[width=0.115\linewidth]{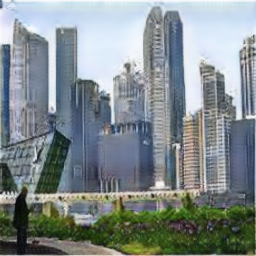}
   \includegraphics[width=0.115\linewidth]{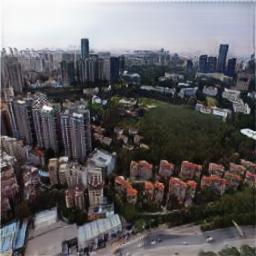}

   \caption{Qualitative comparison of dehazing on real-world images with state-of-the-art methods.}
\end{figure*}

The method of He {\em et al.} \cite{he2011single} is not applicable to the sky regions, as the sky regions are created with a lot of noise. The results generated by the methods of Berman {\em et al.} \cite{berman2016non}, Cai {\em et al.} \cite{cai2016dehazenet} and Ren {\em et al.} \cite{ren2016single} and Li {\em et al.}\cite{li2017aod} have little noise in the sky. But the haze still exists after dehazing and the images are relatively dull. The sky is still dim in dehaze images generated by Pix2pixGAN for haze images with heavy haze. The results of Zhu {\em et al.} \cite{zhu2017haze} have clear sky but not the sunny sky, and haze still exists. The method of Zhang {\em et al.} \cite{zhang2018densely} could produce dehaze images with relatively bright sky. However, the dehaze images still look like captured in an overcast day. Our method could generate clean dehaze images with blue sky. Our model could ensure the visibility of images and restore the scene to have a clean and blue sky. Although the results of CycleGAN have a blue sky, the profile of the objects are fake and messy. The last column image is shotted by our Unmanned Aerial Vehicle (UAV) which is not preprocessed to denoise the image. Its dehaze result by our method is clean with a thin blue sky, in which the most distant building has a clear outline and the beige road in the foreground region does not produce color-shift.

\begin{table*}[htbp]
\begin{center}
\caption{OBJECTIVE ASSESSMENT RESULT.}
\begin{tabular}{|l|c|c|c|c|c|c|c|c|c|c|}
\hline\hline
Method & He & Berman & Cai & Ren & Zhu & Li & Zhang & PixGAN & CycGAN & Our\\
\hline
PSNR(Paired database) & 23.06 & 20.48 & 21.04 & 23.36 & 20.86 & 21.51 & 13.64 & 26.59 & 21.05 & \textbf{27.52}\\
SSIM(Paired database) & 0.81 & 0.73 & 0.66 & \textbf{0.82} & 0.73 & 0.80 & 0.15 & 0.74 & 0.55 & \textbf{0.82}\\
AJAC(Real-world images) & 0.41 & 0.36 &  0.35 & 0.35 & 0.38 & 0.30 & 0.35 & 0.32 & \textbf{0.43} & \textbf{0.43}\\
Fast RCNN(Real-world images) & 33 & 33 & 30 & 30 & 28 & 29 & 32 & 6 & 8 & \textbf{36}\\
\hline
\end{tabular}
\end{center}
\end{table*}
\begin{figure*}[!]
\begin{center}
   \includegraphics[width=0.076\linewidth]{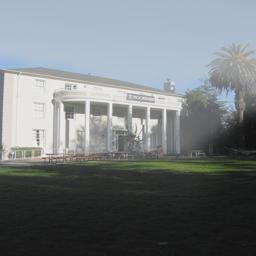}
   \includegraphics[width=0.076\linewidth]{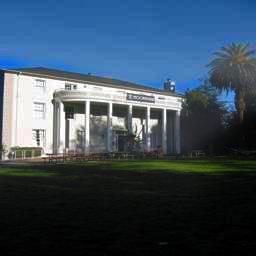}
   \includegraphics[width=0.076\linewidth]{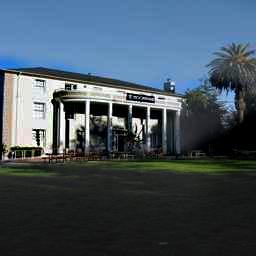}
   \includegraphics[width=0.076\linewidth]{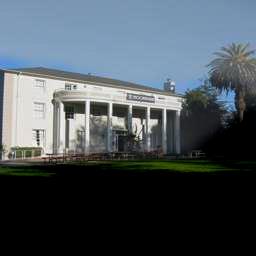}
   \includegraphics[width=0.076\linewidth]{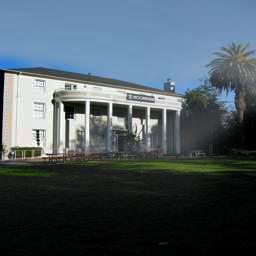}
   \includegraphics[width=0.076\linewidth]{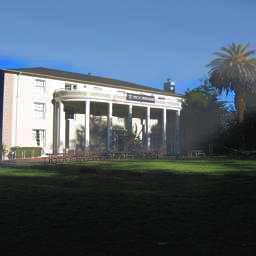}
   \includegraphics[width=0.076\linewidth]{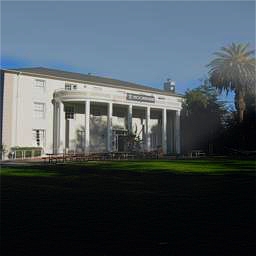}
   \includegraphics[width=0.076\linewidth]{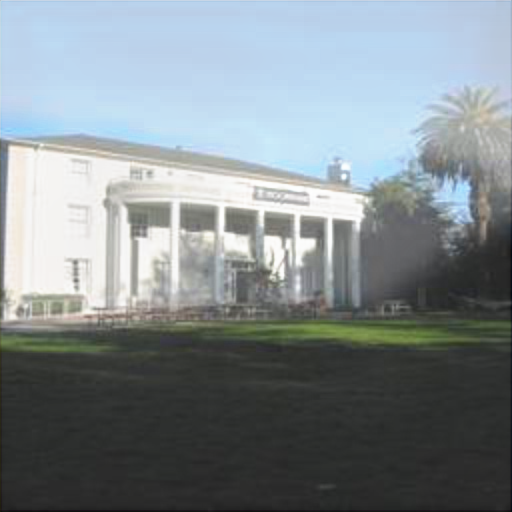}
   \includegraphics[width=0.076\linewidth]{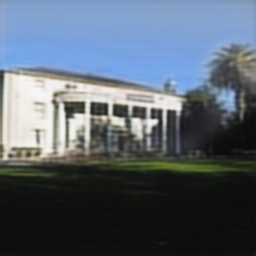}
   \includegraphics[width=0.076\linewidth]{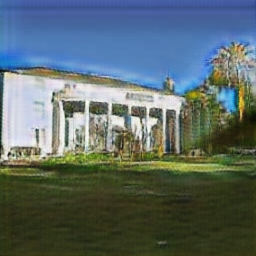}
   \includegraphics[width=0.076\linewidth]{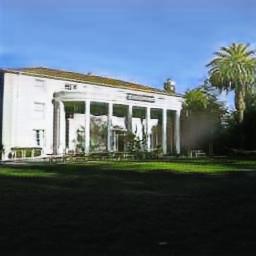}
   \includegraphics[width=0.076\linewidth]{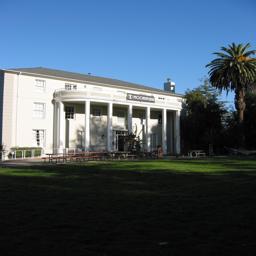}
   \\
   \includegraphics[width=0.076\linewidth]{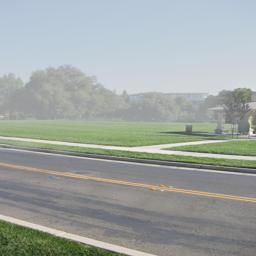}
   \includegraphics[width=0.076\linewidth]{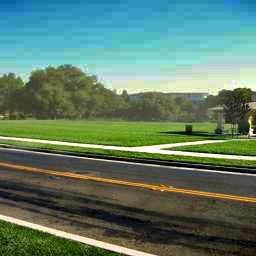}
   \includegraphics[width=0.076\linewidth]{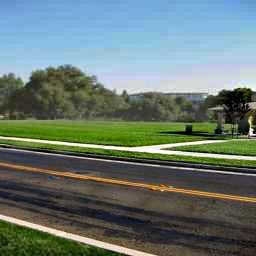}
   \includegraphics[width=0.076\linewidth]{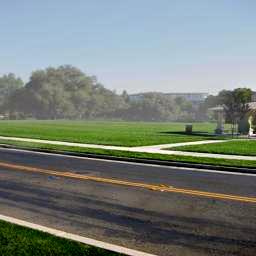}
   \includegraphics[width=0.076\linewidth]{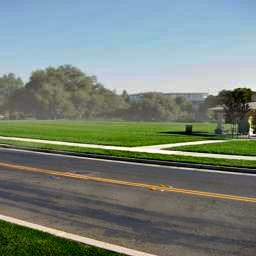}
   \includegraphics[width=0.076\linewidth]{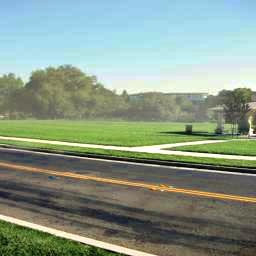}
   \includegraphics[width=0.076\linewidth]{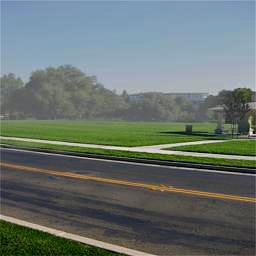}
   \includegraphics[width=0.076\linewidth]{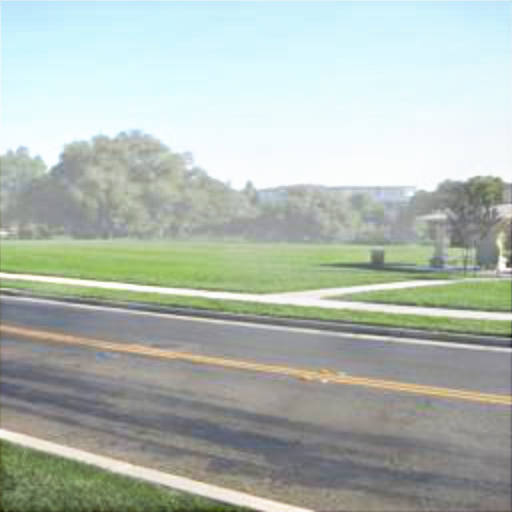}
   \includegraphics[width=0.076\linewidth]{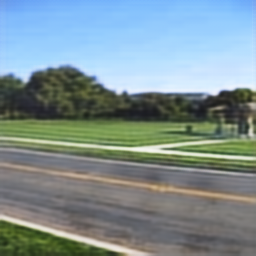}
   \includegraphics[width=0.076\linewidth]{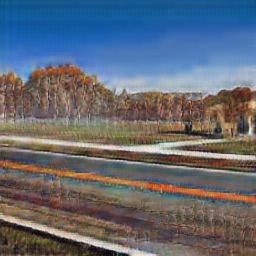}
   \includegraphics[width=0.076\linewidth]{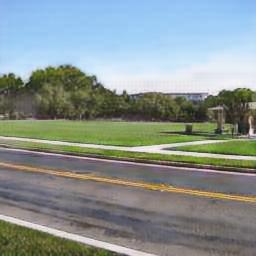}
   \includegraphics[width=0.076\linewidth]{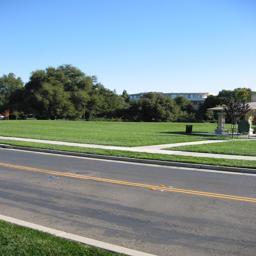}
   \\
   \subfigure[]{\includegraphics[width=0.076\linewidth]{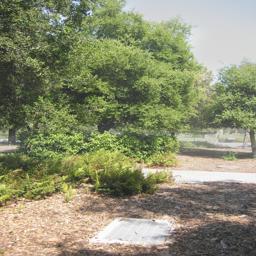}}
   \subfigure[]{\includegraphics[width=0.076\linewidth]{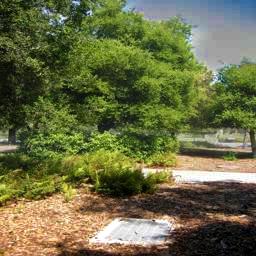}}
   \subfigure[]{\includegraphics[width=0.076\linewidth]{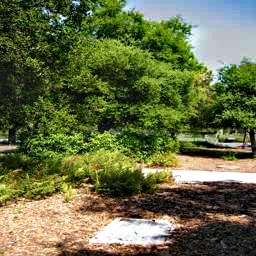}}
   \subfigure[]{\includegraphics[width=0.076\linewidth]{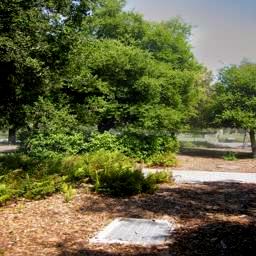}}
   \subfigure[]{\includegraphics[width=0.076\linewidth]{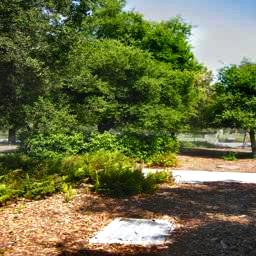}}
   \subfigure[]{\includegraphics[width=0.076\linewidth]{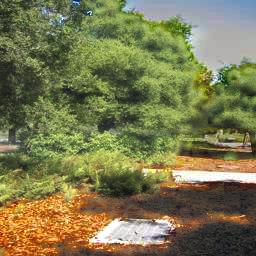}}
   \subfigure[]{\includegraphics[width=0.076\linewidth]{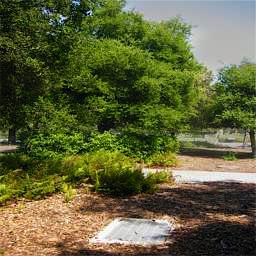}}
   \subfigure[]{\includegraphics[width=0.076\linewidth]{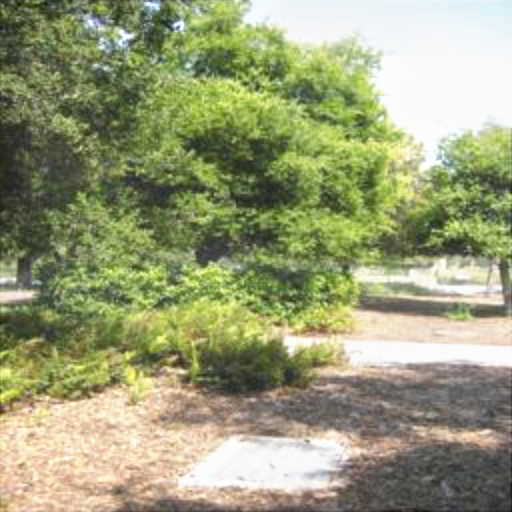}}
   \subfigure[]{\includegraphics[width=0.076\linewidth]{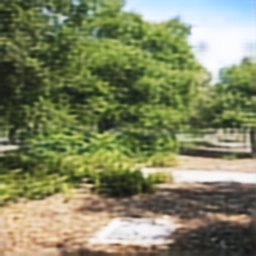}}
   \subfigure[]{\includegraphics[width=0.076\linewidth]{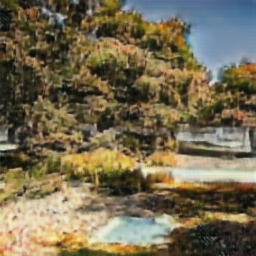}}
   \subfigure[]{\includegraphics[width=0.076\linewidth]{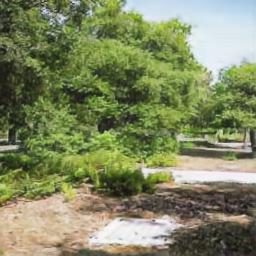}}
   \subfigure[]{\includegraphics[width=0.076\linewidth]{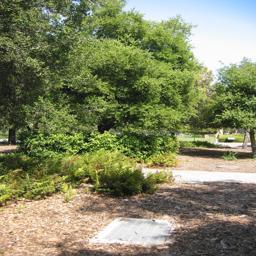}}
   \\
\end{center}
\caption{Qualitative comparison of dehazing on images with state-of-the-art methods. (a)The original images. (b)The results of He {\em et al.} \cite{he2011single}. (c)The results of Berman {\em et al.} \cite{berman2016non}. (d)The results of Cai {\em et al.}. \cite{cai2016dehazenet}. (e)The results of Ren {\em et al.} \cite{ren2016single}. (f) The results of Zhu {\em et al.} \cite{zhu2017haze}.  (g)The results of Li {\em et al.} \cite{li2017aod}. (h)The results of Zhang {\em et al.} \cite{zhang2018densely}. (i)The results of Pix2pixGAN \cite{isola2016image}. (j)The results of CycleGAN \cite{zhu2017unpaired}. (k)The results of our SCGAN. (l)The groundtruth.}
\end{figure*}
\begin{figure*}[!]
\begin{center}
   \subfigure[2]{\includegraphics[width=0.085\linewidth]{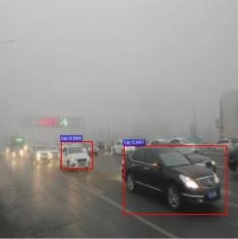}}
   \subfigure[2]{\includegraphics[width=0.085\linewidth]{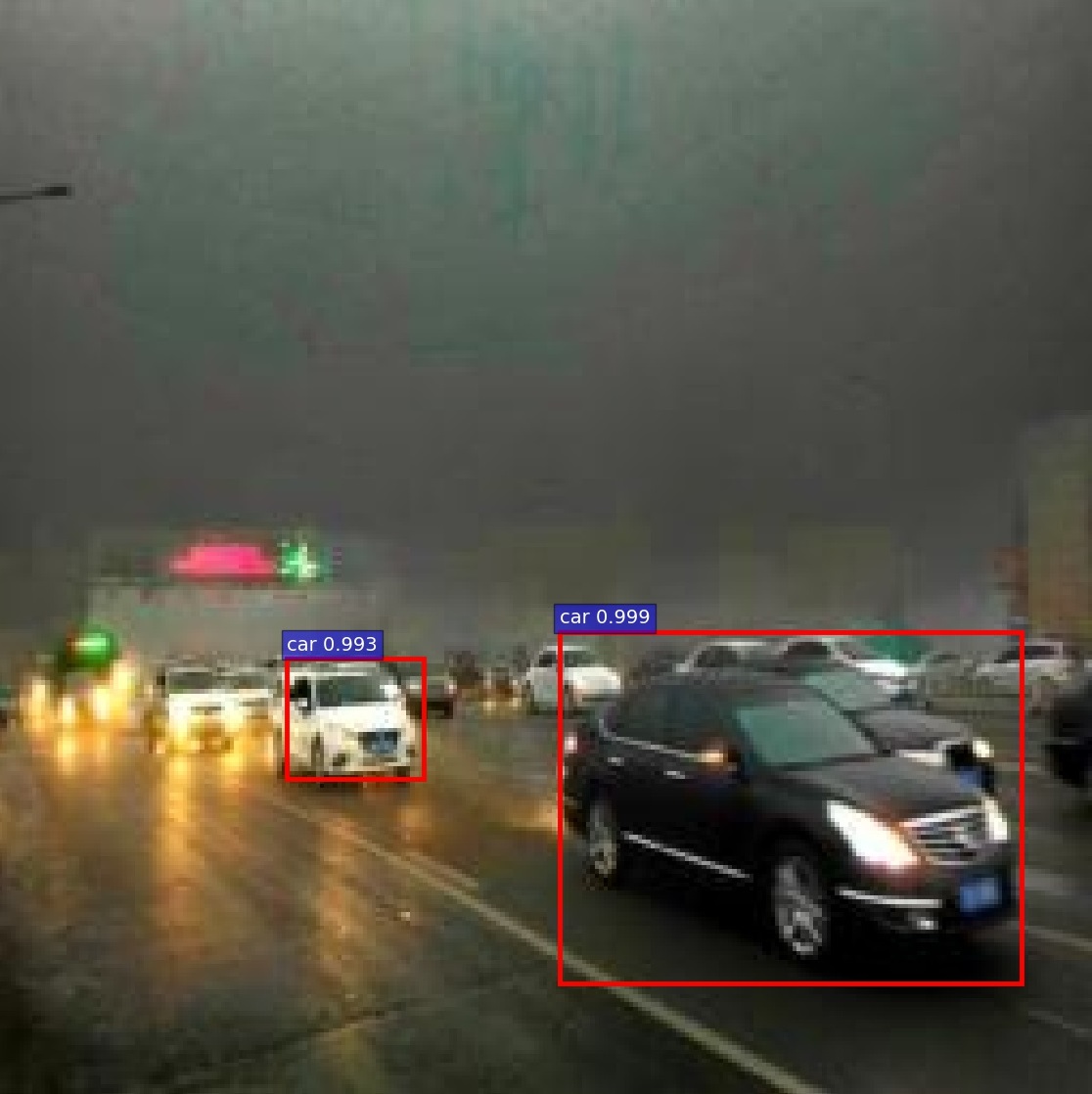}}
   \subfigure[2]{\includegraphics[width=0.085\linewidth]{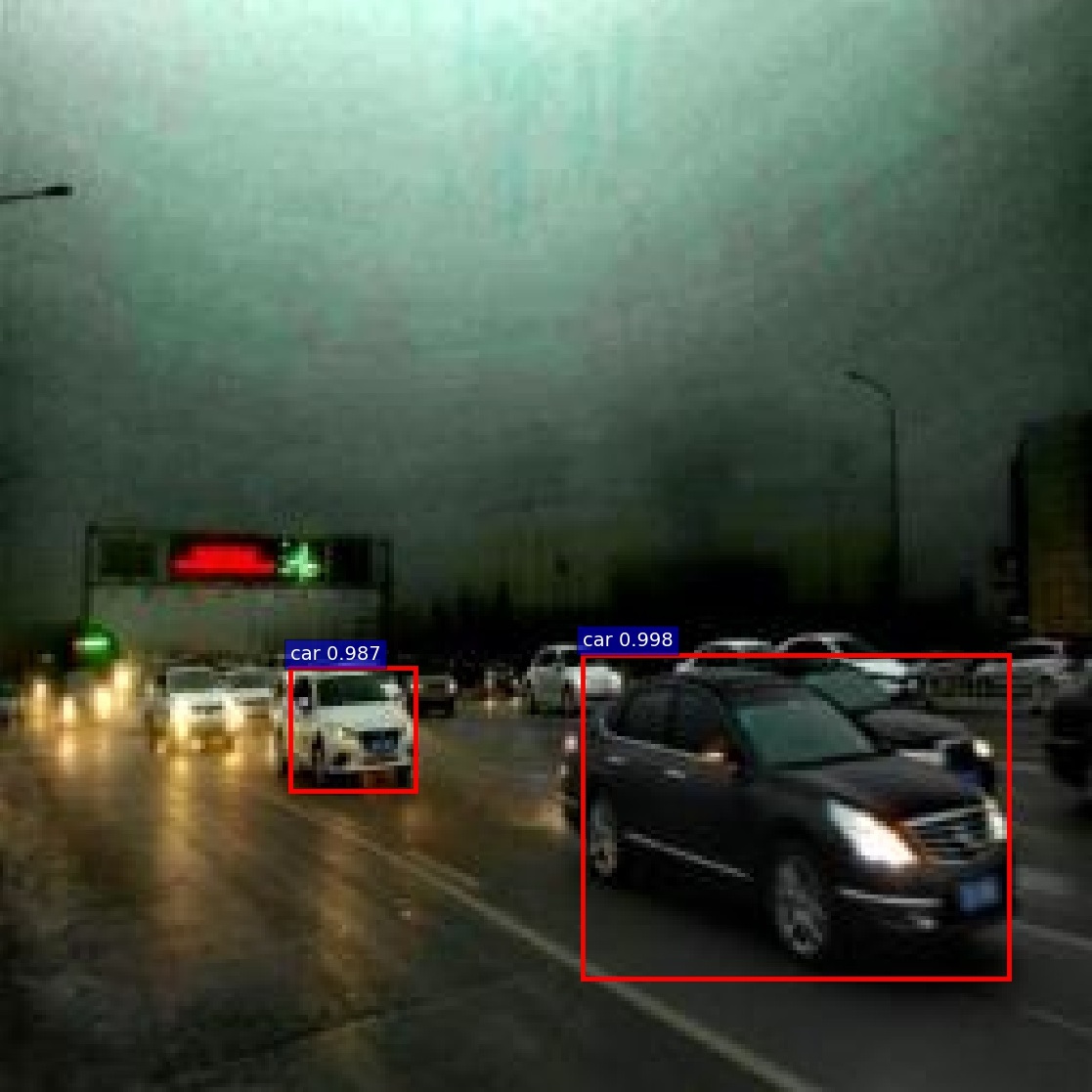}}
   \subfigure[2]{\includegraphics[width=0.085\linewidth]{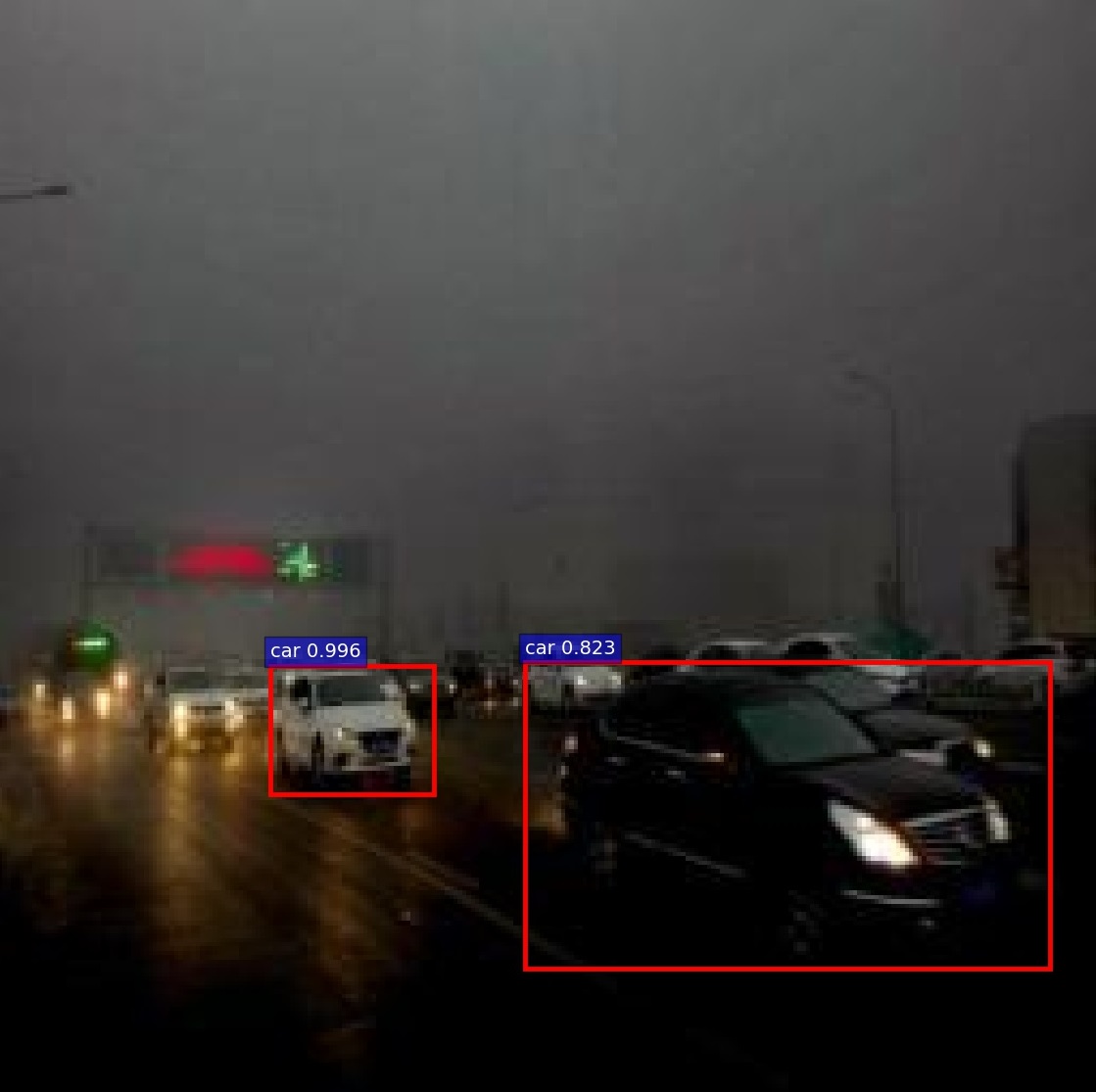}}
   \subfigure[2]{\includegraphics[width=0.085\linewidth]{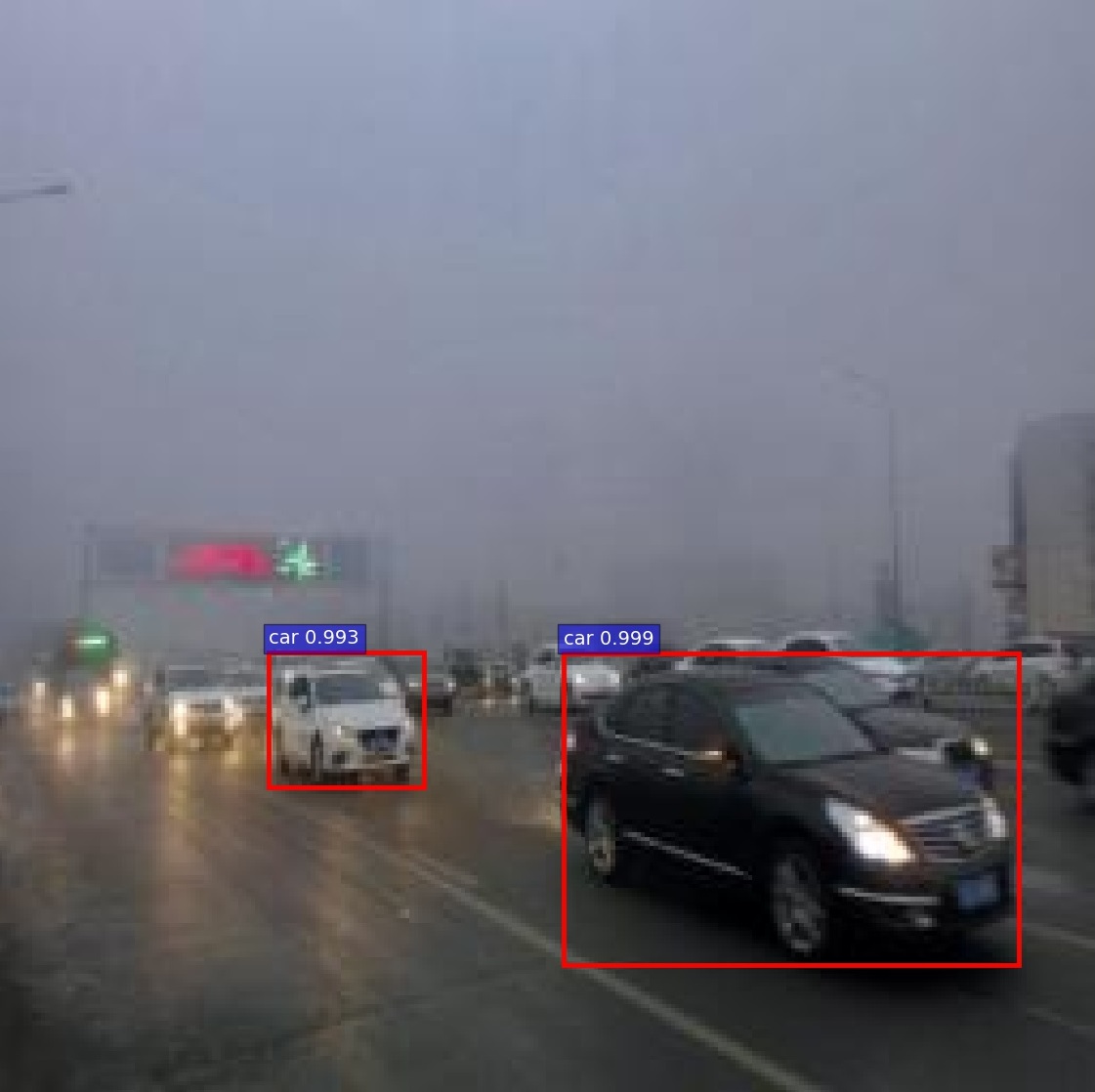}}
   \subfigure[2]{\includegraphics[width=0.085\linewidth]{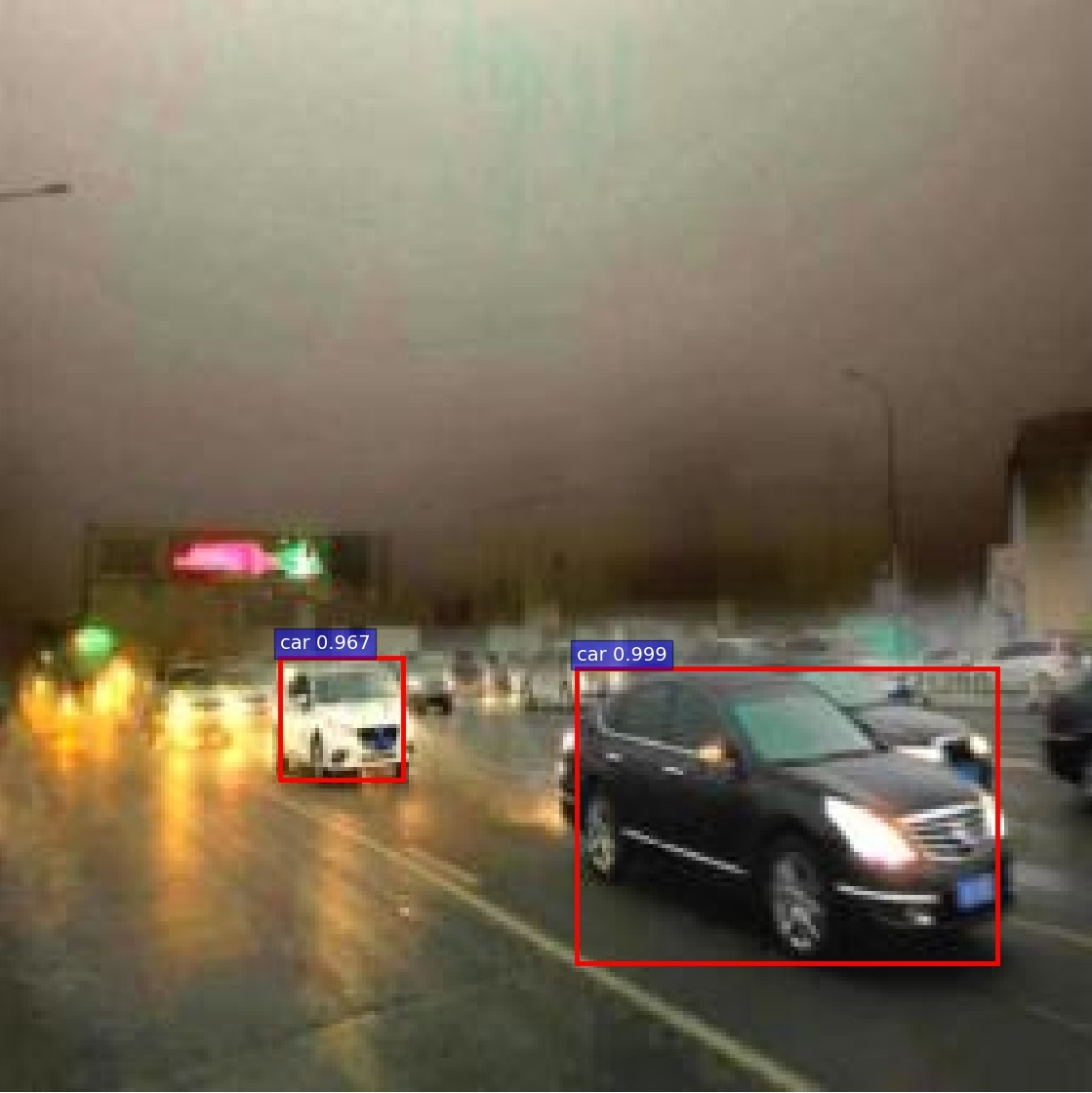}}
   \subfigure[1]{\includegraphics[width=0.085\linewidth]{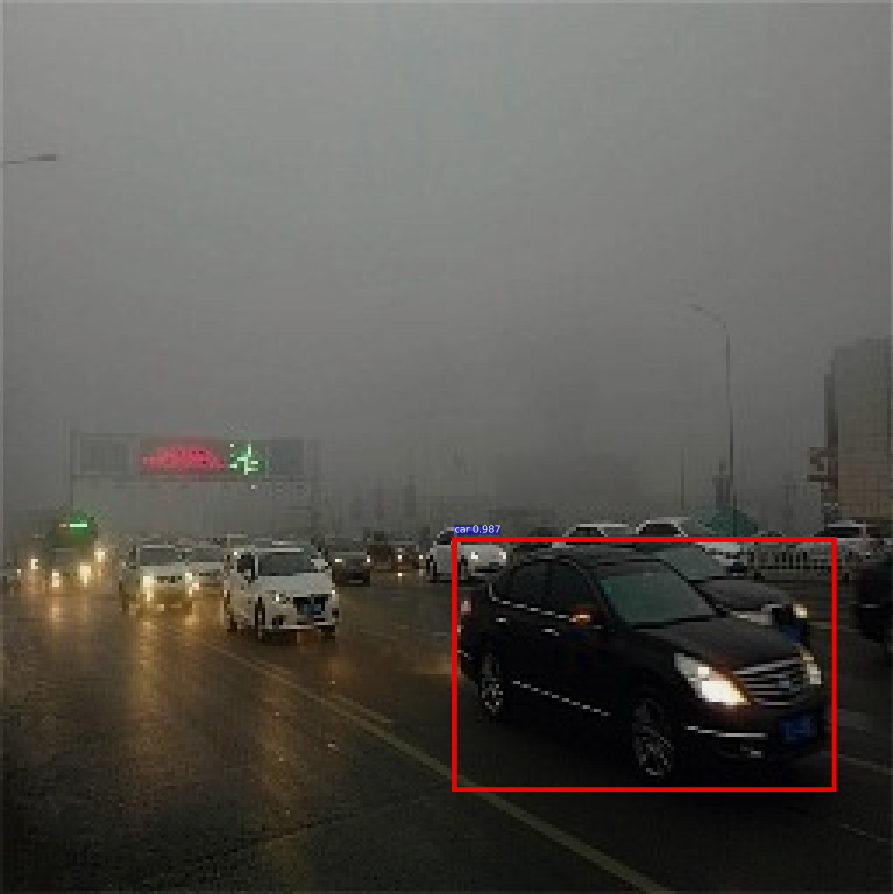}}
   \subfigure[2]{\includegraphics[width=0.085\linewidth]{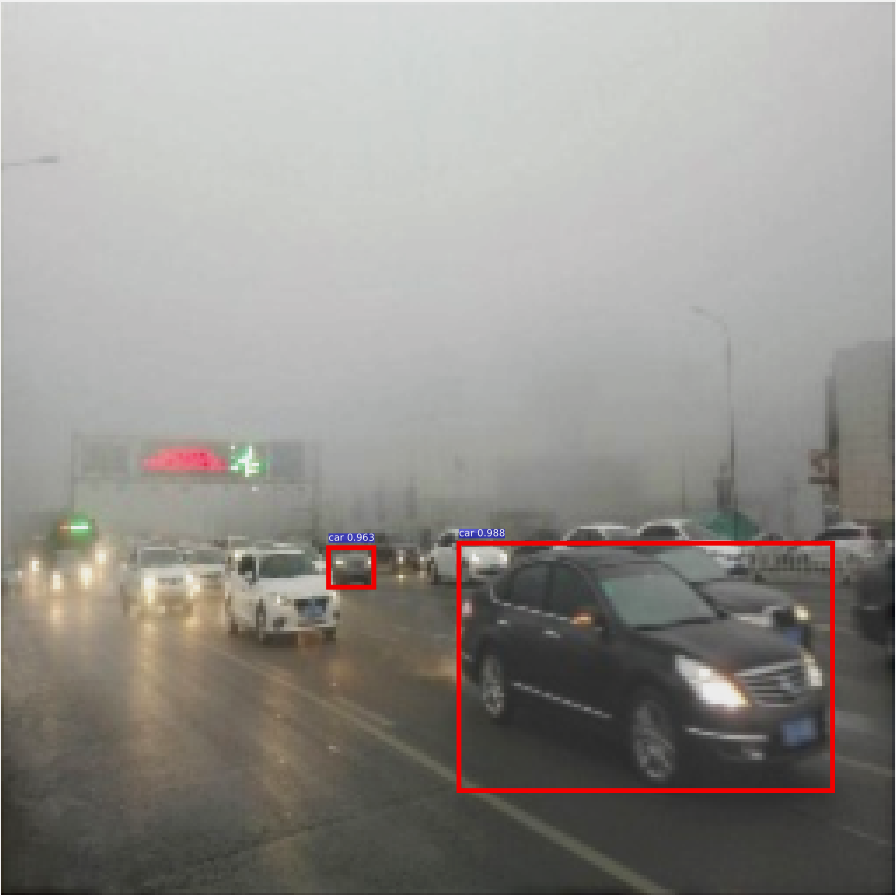}}
   \subfigure[1]{\includegraphics[width=0.085\linewidth]{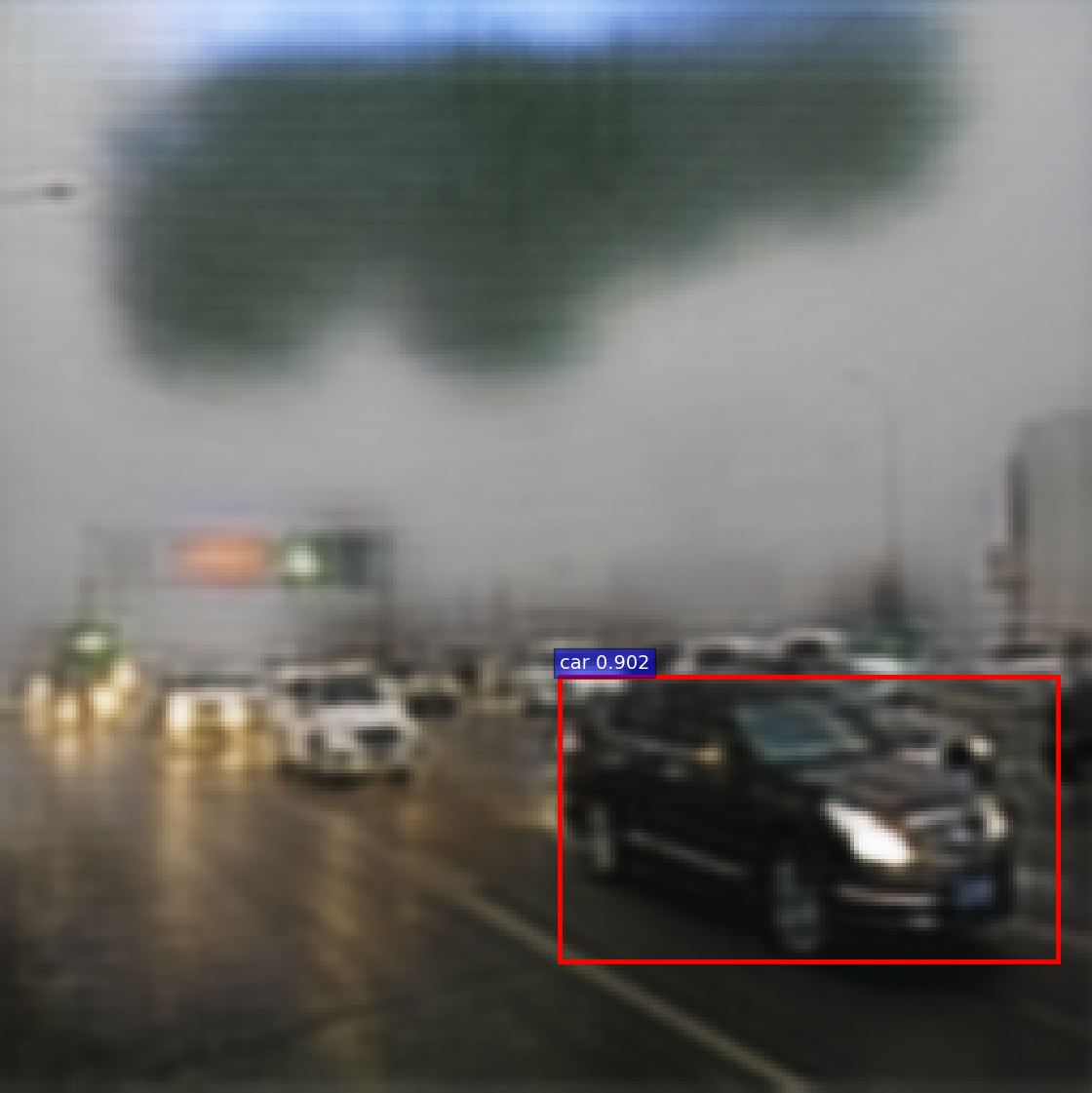}}
   \subfigure[0]{\includegraphics[width=0.085\linewidth]{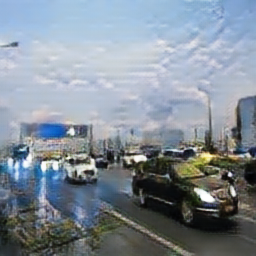}}
   \subfigure[3]{\includegraphics[width=0.085\linewidth]{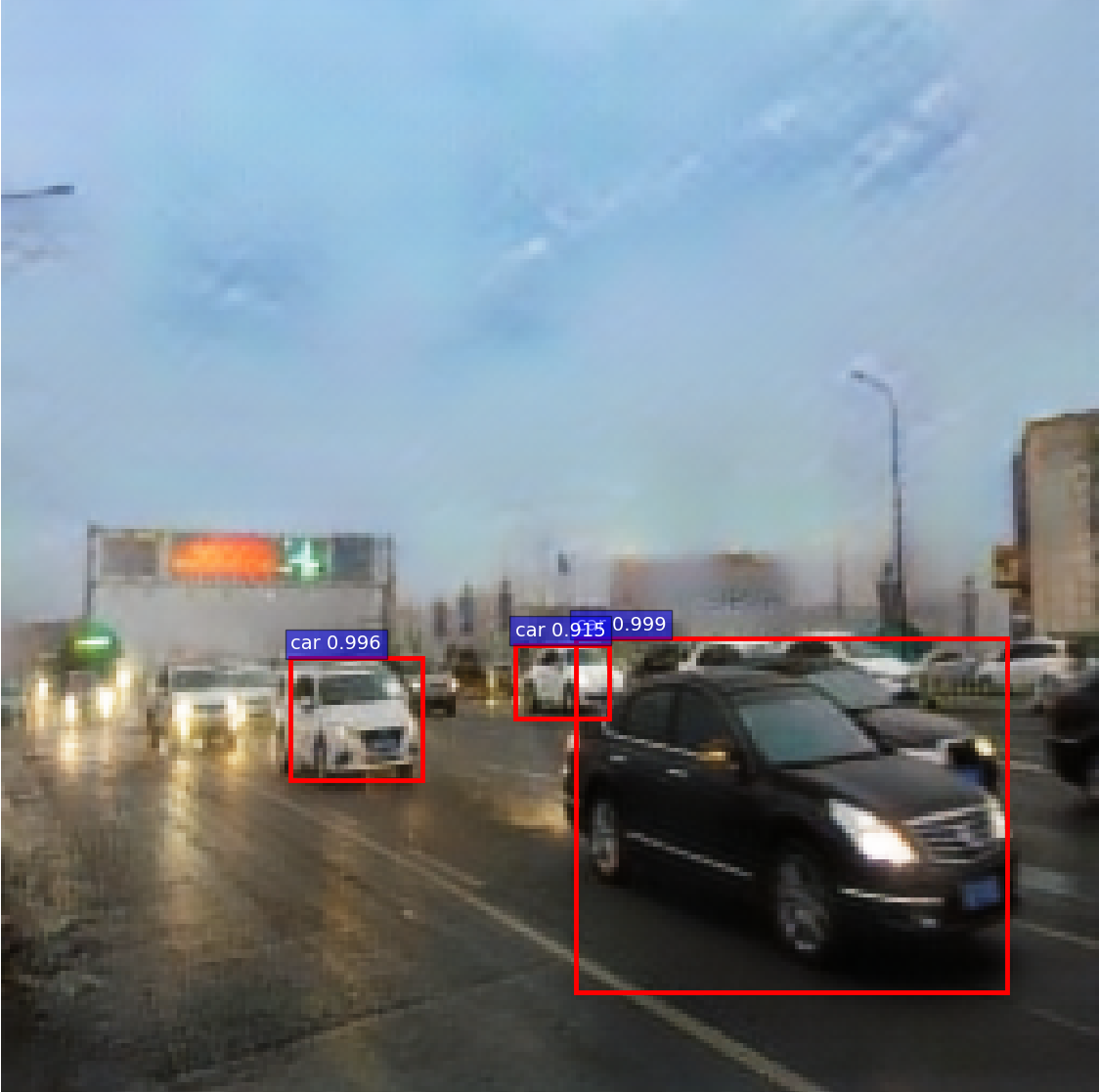}}
\end{center}
\caption{Car detection comparison of dehazing on real-world images with state-of-the-art methods. The caption of each result is the number of vehicles detected. (a)The original image. (b)The result of He {\em et al.} \cite{he2011single}. (c)The result of Berman {\em et al.} \cite{berman2016non}. (d)The result of Cai {\em et al.} \cite{cai2016dehazenet}. (e)The result of Ren {\em et al.} \cite{ren2016single}. (f)The result of Zhu {\em et al.} \cite{zhu2017haze}. (g)The result of Li {\em et al.} \cite{li2017aod}. (h)The result of Zhang {\em et al.} \cite{zhang2018densely}. (i)The result of Pix2pixGAN \cite{isola2016image}. (j)The result of CycleGAN {\em et al.} \cite{zhu2017unpaired}. (k)The result of our SCGAN.}
\end{figure*}
\subsection{Objective Assessment}
We also analyze the quantitative evaluation metrics to quantitatively assess our proposed algorithm. Firstly we evaluate the haze removal efficiency by 100 paired images. The haze images are generated by adding haze on the haze-free images which include depth information. The test dataset is collected from Make3d dataset \cite{saxena20083,saxena2009make3d}. The way of adding haze is defined on Section \ref{subsec:data}. PSNR and SSIM are the criteria in terms of the difference between each pair of the ground truth haze-free image and dehaze result. As can be seen from Table \uppercase\expandafter{\romannumeral3}, our results almost have the highest PSNR and SSIM values comparing to other algorithms. They reflect that our dehaze images are of the highest quality, and closest to the ground truth. Ren {\em et al.} \cite{ren2016single} has similar SSIM value with our method. The reason is that their results preserve the original image structures with less artifacts. However, the PSNR value of Ren {\em et al.} \cite{ren2016single} is lower, because their results still have haze in the sky region.

The dehaze examples of state-of-the-art methods and our method on several haze images are compared to the ground truth in Fig. 9. Our results in Fig. 9(k) are the closest to the ground truth in Fig. 9(l). The white house in the first example figure is dim in the results of most methods. In the second example, the road in the results is over-processed except in our method. In the third example, the path fulled of leaves is over-processed in most of results. But our method is not over-processed at the green leaves and the sky region is avoided from color-shift. Our results are close to the ground truth images.

Our method is to restore clear images, which is to let the image has a better visual sense after dehazing. So the aesthetic appreciation value of the image will also be improved after dehazing. Therefore, we test the aesthetic score by using a rank photo aesthetics method which considers meaningful photographic attributes and image content information (AJAC) \cite{kong2016photo}. We test 300 real-world haze images by using the ten haze removal methods and compute the average AJAC score for each method. The average AJAC score of our method is higher than those of other methods (except CycleGAN, see Table \uppercase\expandafter{\romannumeral3}). This conveys that the dehaze results generated by our method have higher aesthetic appearance than those generated by traditional image dehaze methods. The CycleGAN has the same aesthetic score with our method as it also generates colorful images with blue sky. The aesthetic score of He {\em et al.} \cite{he2011single} is quite close to our score. Because their results have high contrast which is preferred in aesthetic measurement.

We also collect 30 haze images with the road contained vehicles to test our haze removal effect. These images are first dehazed using haze removal methods and then car detection is performed. A better haze removal effect should have chance to detect more cars. Fast-RCNN \cite{ren2017faster} is used to detect cars. The official checkpoints of Fast RCNN is trained on Pascal-VOC 2007 dataset. We test the above ten different methods. The number of vehicles detected by our method is the highest (Table \uppercase\expandafter{\romannumeral3}). This fully demonstrates that the dehaze images by our method is the clearest. An example of the detection comparison is shown in Fig. 10. Our result has a blue sky instead of a dark sky and the foreground is fresh and bright. Three cars are detected in our dehaze image while only 2 are detected in results of most other methods.

\subsection{Assessment on Public Dataset}
\begin{table*}[htbp]
\begin{center}
\caption{Assessment on Public Datasets of RESIDE HSTS and O-haze}
\begin{tabular}{|l|c|c|c|c|c|c|c|c|c|c|}
\hline\hline
Method & He & Berman & Cai & Ren & Zhu & Li & Zhang & Our\\
\hline
PSNR(HSTS) & 19.04 & 19.43 & \textbf{25.76} & 19.35 & 21.79 & 20.55 & 21.81 & 21.10\\
SSIM(HSTS) & 0.75 & 0.77 & \textbf{0.88} & 0.76 & 0.79 & 0.78 & 0.67 & 0.78\\
AJAC(HSTS) & \textbf{0.49} & 0.48 & 0.44 & 0.46 & 0.43 & 0.44 & 0.42 & \textbf{0.49}\\
\hline
PSNR(O-haze) & 15.75 & 16.21 & 17.14 & 17.54 & 15.16 & \textbf{17.76} & 17.60 & 17.45\\
SSIM(O-haze) & 0.46 & \textbf{0.55} & 0.44 & 0.52 & 0.45 & 0.42 & 0.38 & 0.53\\
AJAC(O-haze) & 0.50 & 0.47 & 0.39 & 0.43 & 0.47 & 0.36 & 0.38 & \textbf{0.56}\\
\hline
\end{tabular}
\end{center}
\end{table*}

\begin{figure*}[!]
\label{fig:11}
\begin{center}
   \includegraphics[width=0.085\linewidth]{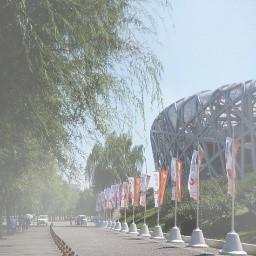}
   \includegraphics[width=0.085\linewidth]{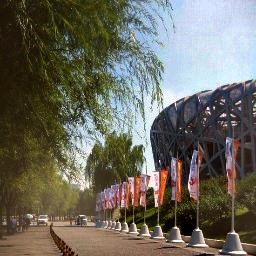}
   \includegraphics[width=0.085\linewidth]{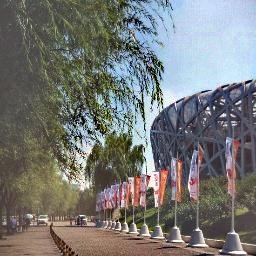}
   \includegraphics[width=0.085\linewidth]{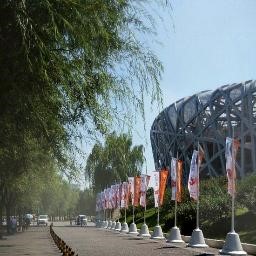}
   \includegraphics[width=0.085\linewidth]{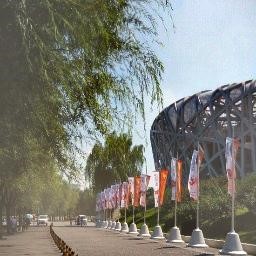}
   \includegraphics[width=0.085\linewidth]{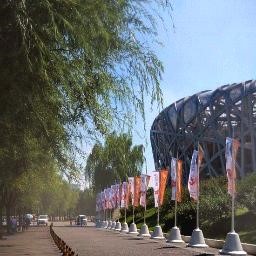}
   \includegraphics[width=0.085\linewidth]{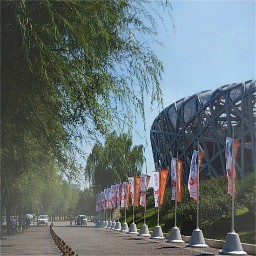}
   \includegraphics[width=0.085\linewidth]{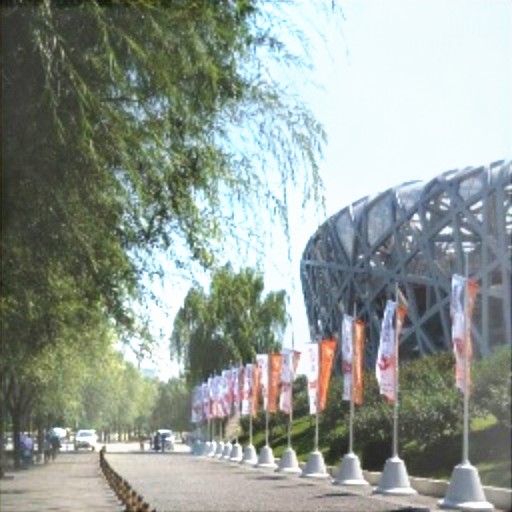}
   \includegraphics[width=0.085\linewidth]{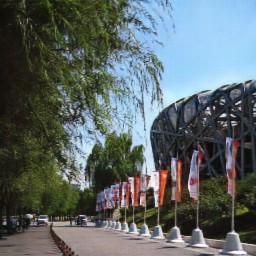}
   \includegraphics[width=0.085\linewidth]{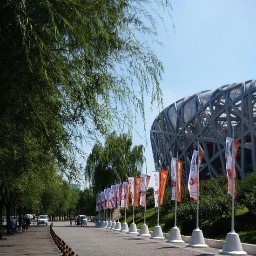}\\
   \subfigure[]{\includegraphics[width=0.085\linewidth]{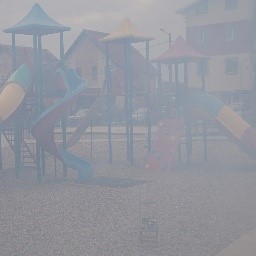}}
   \subfigure[]{\includegraphics[width=0.085\linewidth]{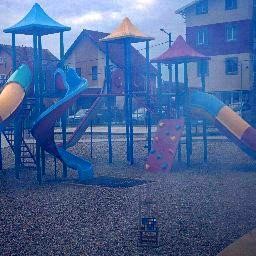}}
   \subfigure[]{\includegraphics[width=0.085\linewidth]{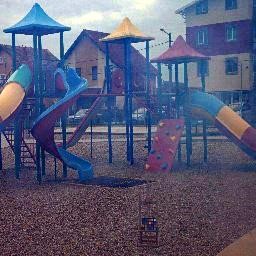}}
   \subfigure[]{\includegraphics[width=0.085\linewidth]{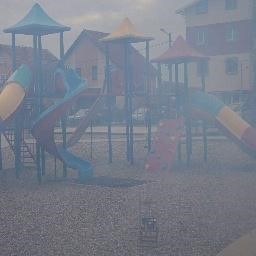}}
   \subfigure[]{\includegraphics[width=0.085\linewidth]{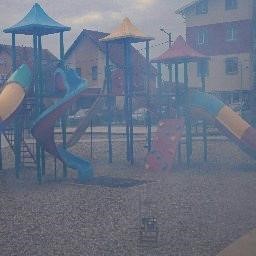}}
   \subfigure[]{\includegraphics[width=0.085\linewidth]{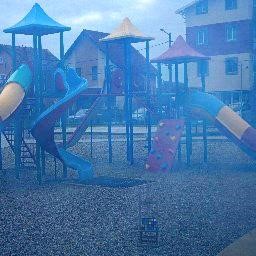}}
   \subfigure[]{\includegraphics[width=0.085\linewidth]{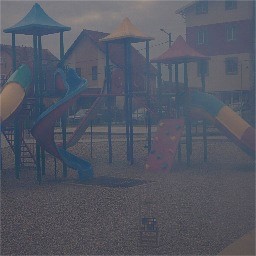}}
   \subfigure[]{\includegraphics[width=0.085\linewidth]{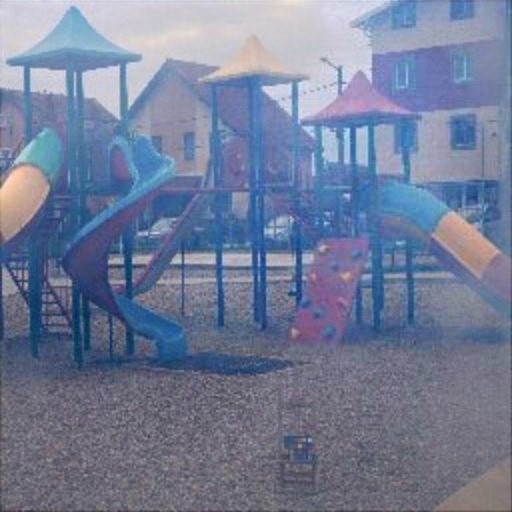}}
   \subfigure[]{\includegraphics[width=0.085\linewidth]{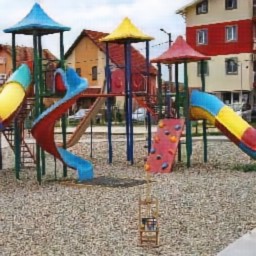}}
   \subfigure[]{\includegraphics[width=0.085\linewidth]{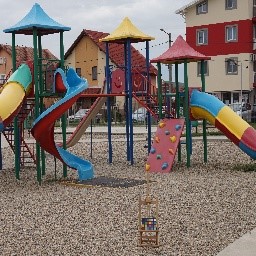}}
\end{center}
\caption{Dehaze results of images from public datasets. The image in the first row is from RESIDE HSTS, the image in the second row is from O-haze. (a)The original images. (b)The results of He {\em et al.} \cite{he2011single}. (f)The results of Berman {\em et al.} \cite{berman2016non}. (c)The results of Cai {\em et al.} \cite{cai2016dehazenet}. (e)The results of Ren {\em et al.} \cite{ren2016single}. (d)The results of Zhu {\em et al.} \cite{zhu2017haze}. (g)The results of Li {\em et al.} \cite{li2017aod}. (h)The results of Zhang {\em et al.} \cite{zhang2018densely}. (i)The results of our SCGAN. (j)The groundtruth.}
\end{figure*}

To conduct fair comparison with related methods, we compared the performance of our model with related methods on public outdoor dahaze datasets RESIDE HSTS (Hybrid Subjective Testing Set) \cite{Boyi2019} and O-haze \cite{Ancuti2018}. The RESIDE HSTS dataset contains 10 synthetic outdoor hazy images with corresponding ground truth haze-free images. The O-HAZE dataset contains 45 different outdoor scenes depicting the same visual content recorded in haze-free and hazy conditions, under the same illumination parameters. We tested the related seven state-of-the-art methods and our model on these two datasets for haze removal. We measured the PSNR, SSIM and AJAC \cite{kong2016photo} scores of these methods on the two datasets, which are summarized in Table \uppercase\expandafter{\romannumeral4}. As our method generate dehaze images in size of $256 \times 256$, therefore, we resize all haze images to $256 \times 256$ in the testing. It may cause that the scores we get are different from those presented in \cite{Boyi2019,Ancuti2018}. In addition, the SSIM values of the related methods on O-haze dataset are obvious different from those reported in \cite{Ancuti2018}, because we did not normalize the image before calculating SSIM.

Table \uppercase\expandafter{\romannumeral4} shows the PSNR, SSIM and AJAC \cite{kong2016photo} scores on the two public test datasets. Our method has comparable or better PSNR and SSIM values on RESIDE HSTS than the methods of He {\em et al.} \cite{he2011single}, Berman {\em et al.} \cite{berman2016non}, Ren {\em et al.} \cite{ren2016single}, Zhu {\em et al.} \cite{zhu2017haze}, Li {\em et al.}\cite{li2017aod} and Zhang {\em et al.} \cite{zhang2018densely}. The results of Cai {\em et al.} \cite{cai2016dehazenet} have the highest PSNR and SSIM values on RESIDE HSTS. It shows that the details and profile of dehaze results by Cai {\em et al.} \cite{cai2016dehazenet} are more similar with grounthtruth than those of our model. However, the dehaze results of our model have the highest aesthetic score, as our model could generate dehaze images with appeal sky, as the example show in first row of Fig. 11. On the dataset of O-haze, our model also has comparable PSNR and SSIM values comparing with the seven related methods, while has the highest aesthetic score. The results of Berman {\em et al.} \cite{berman2016non} have relatively clear scene but have heavy color-shift in O-haze images (see the example in the second row of Fig.11). Comparing with results of the related methods (see Fig.11), our model could successfully remove the haze of the images. But the color of the images, especially the sky region is biased to bright blue, which is different from the ground truth. This is one of the reason that the PSNR of our SCGAN model is not the highest comparing with other methods. In summary, the assessment on the public datasets shows that our model could generate more visual aesthetic dehaze images while having comparable or better dehaze ability comparing with related state-of-the-art dehaze methods.

\section{Limitation}
Our limitation is that some of the details are still unrealistic in the dehaze results, because the supervised model relies on training dataset, and we only have limited number of paired images and the scene could not cover all possible content. Additionally, our method may change the color of the original sky to fit a blue sky with white cloud. The scene in sunset is changed to high noon with blue sky in the first example of Fig. 12. The reason is that most of the haze-free images in our unpaired dataset have a blue sky with or without distributed white cloud.

Our method could also remove the haze of images without sky, as the examples show in the second and third examples of Fig. 12. However, regions with heavy haze may be considered as a sky region that are reconstructed to be blue.
\begin{figure}[t!]
\begin{center}
    \includegraphics[width=0.21\linewidth]{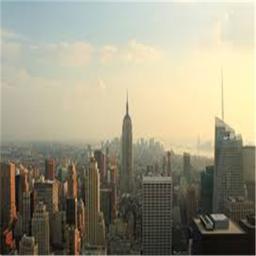}
    \includegraphics[width=0.21\linewidth]{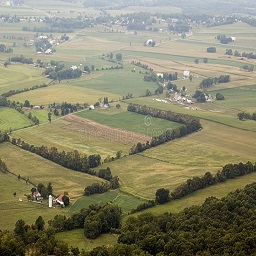}
    \includegraphics[width=0.21\linewidth]{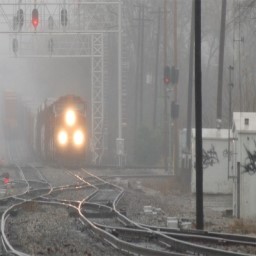}\\
    \includegraphics[width=0.21\linewidth]{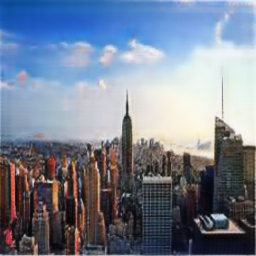}
    \includegraphics[width=0.21\linewidth]{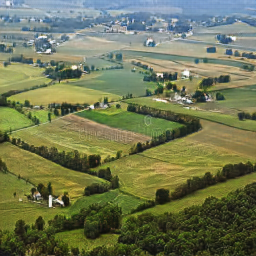}
    \includegraphics[width=0.21\linewidth]{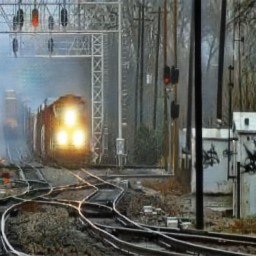}
   \end{center}
   \caption{The limitation of our method. Top: Original image. The second row: Dehaze result using our SCGAN model.}
\end{figure}

\section{Conclusion}
This paper propose a novel end-to-end dehaze model with cycle structure which includes two generators and two discriminators. To effectively solve the lack of training data, we use two different outdoor real-world datasets to train our model: paired image dataset and an unpaired image dataset. Four different kinds of loss function are used to constrain the effect including adversarial loss, cycle consistency loss, photorealism loss and paired L1 loss. The photorealism loss helps our model successfully refine the edges of images to close to those of the real images. Our model is effective on image dehazing and also could reconstruct the sky to be clean and blue (likely captured in a sunny weather). The experiments have show that our model could generate more visual aesthetic dehaze images while having comparable or better dehaze ability comparing with related state-of-the-art dehaze methods.

\section*{Acknowledgement}
\label{sec:9}
This work was supported in part by: (i) National Natural Science Foundation of China (Grant No. 61602314, 61602313, and 61620106008); (ii) Natural Science Foundation of Guangdong Province of China (Grant No. 2016A030313043); (iii) Fundamental Research Project in the Science and Technology Plan of Shenzhen (Grant No. JCYJ20160331114551175).
\bibliography{egbib}
\section*{Appendix}
\label{appendix}

  \section*{Most experimental results}

\begin{figure*}[!]
\begin{center}
   \includegraphics[width=0.2\linewidth]{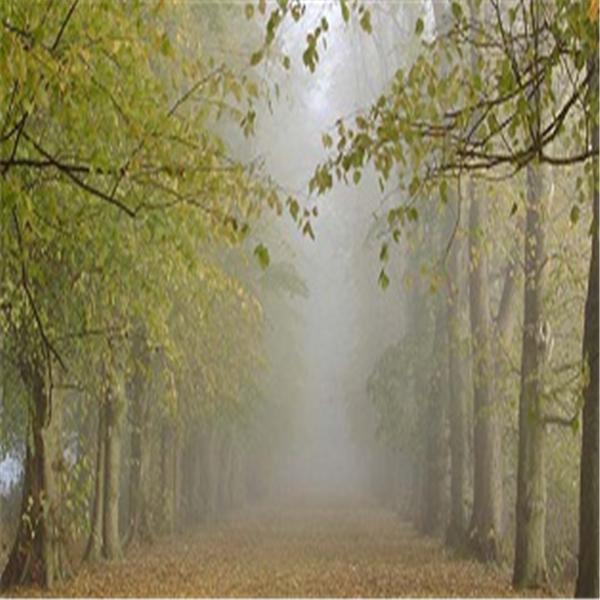}
   \includegraphics[width=0.2\linewidth]{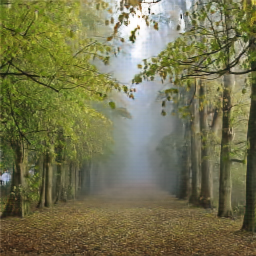}
   \includegraphics[width=0.2\linewidth]{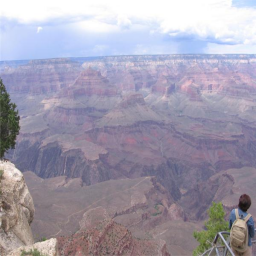}
   \includegraphics[width=0.2\linewidth]{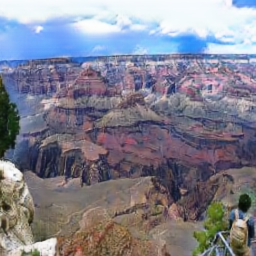}\\
   \includegraphics[width=0.2\linewidth]{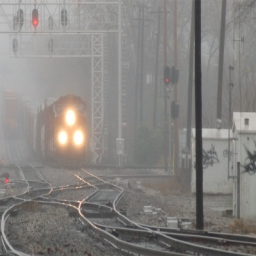}
   \includegraphics[width=0.2\linewidth]{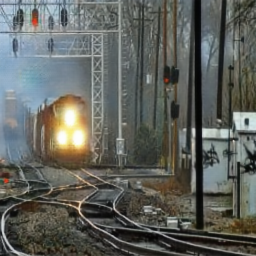}
   \includegraphics[width=0.2\linewidth]{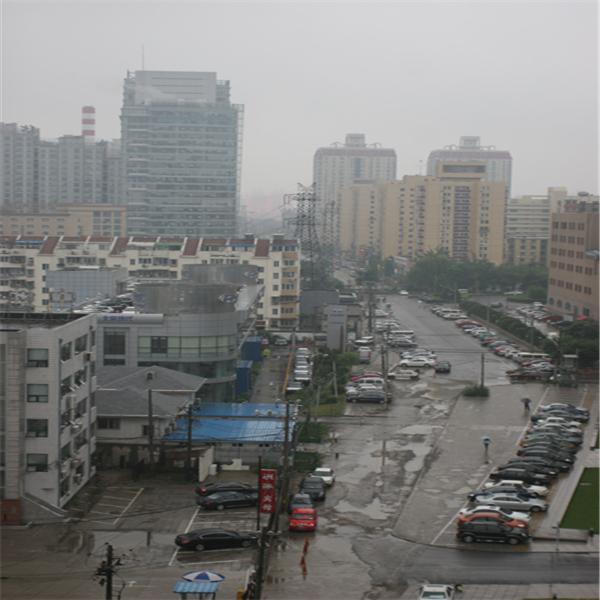}
   \includegraphics[width=0.2\linewidth]{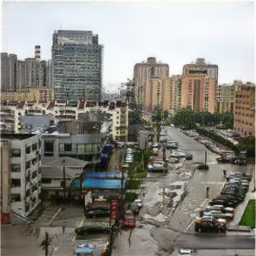}\\
   \includegraphics[width=0.2\linewidth]{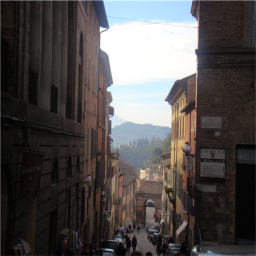}
   \includegraphics[width=0.2\linewidth]{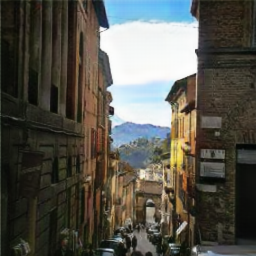}
   \includegraphics[width=0.2\linewidth]{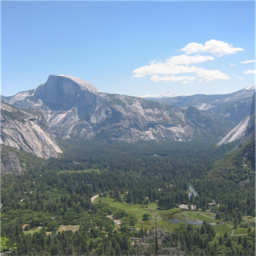}
   \includegraphics[width=0.2\linewidth]{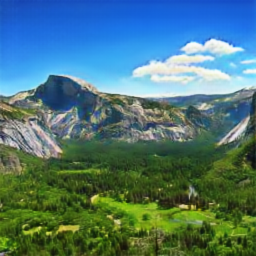}\\
   \includegraphics[width=0.2\linewidth]{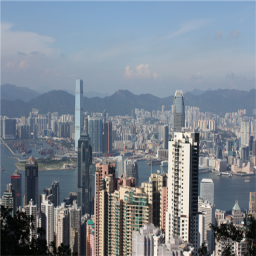}
   \includegraphics[width=0.2\linewidth]{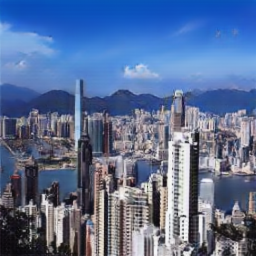}
  \includegraphics[width=0.2\linewidth]{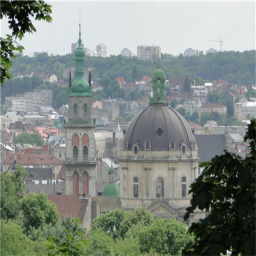}
  \includegraphics[width=0.2\linewidth]{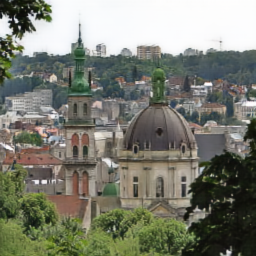}
   \end{center}
   \caption{More experiment results.}
\label{fig:10_1}
\end{figure*}
\end{document}